\documentclass{article} % For LaTeX2e
\usepackage{iclr2019_conference,times}
\usepackage{multirow}
\usepackage{url}
\usepackage{longtable}

\usepackage[paperVersion]{marcpaper}

\title{Learning to Represent Edits}

\iclrfinalcopy

\author{Pengcheng Yin\thanks{Work done as an intern in Microsoft Research, Cambridge, UK.},~~Graham Neubig \\
Language Technology Institute\\
Carnegie Mellon University\\
Pittsburgh, PA 15213, USA \\
\texttt{\{pcyin,gneubig\}@cs.cmu.edu} \\
\AND
Miltiadis Allamanis, Marc Brockschmidt, Alexander L. Gaunt \\
Microsoft Research \\
Cambridge, CB1 2FB, United Kingdom \\
\texttt{\{miallama,mabrocks,algaunt\}@microsoft.com} \\
}

\newcommand{\element}{\ensuremath{\vect{x}}\xspace}
\newcommand{\elementbefore}{\ensuremath{\element_-}\xspace}
\newcommand{\elementafter}{\ensuremath{\element_+}\xspace}
\newcommand{\change}[2]{\ensuremath{#1 \rightarrow #2}\xspace}
\newcommand{\changeReprFunc}{\ensuremath{f_\Delta}\xspace}
\newcommand{\editrepr}[2]{\ensuremath{\changeReprFunc(#1,#2)}\xspace}
\newcommand{\editorSym}{\ensuremath{\alpha}\xspace}
\newcommand{\editorFunc}[2]{\ensuremath{\editorSym(#1,#2)}\xspace}

\newcommand{\expandRule}{\textsc{ExpandR}\xspace}
\newcommand{\applyConstr}{\textsc{ApplyConstr}\xspace}
\newcommand{\reduce}{\textsc{Reduce}\xspace}
\newcommand{\genTerminal}{\textsc{GenTerm}\xspace}
\newcommand{\subtreecopy}{\textsc{TreeCp}\xspace}

\begin{document}

\maketitle

\begin{abstract}
We introduce the problem of learning distributed representations of edits. By combining a
``neural editor'' with an ``edit encoder'', our models learn to represent the salient
information of an edit and can be used to apply edits to new inputs.
We experiment on natural language and source code edit data. Our evaluation yields
promising results that suggest that our neural network models learn to capture
the structure and semantics of edits. We hope that this interesting task and
data source will inspire other researchers to work further on this problem.
\end{abstract}

\section{Introduction}
%!TEX root = ../editrepresentations.tex

One great advantage of electronic storage of documents is the ease
with which we can edit them, and edits are performed in a wide variety
of contents.
For example, right before a conference deadline, papers worldwide are
finalized and polished, often involving common fixes for grammar,
clarity and style. Would it be possible to automatically extract
rules from these common edits?
Similarly, program source code is constantly changed to implement
new features, follow best practices and fix bugs.
With the widespread deployment of (implicit) version control systems,
these edits are quickly archived, creating a major data stream that we can learn
from.

In this work, we study the problem of learning distributed
representations of edits.
We only look at small edits with simple semantics that are more 
likely to appear often and do not consider larger edits; \ie,
we consider ``add definite articles'' rather than ``rewrite act 2,
scene 3.''
Concretely, we focus on two questions:
\begin{enumerate*}[label=\roman*)]
 \item Can we group semantically equivalent edits together, so that
  we can automatically recognize common edit patterns?
 \item Can we automatically transfer edits from one context to
  another?
\end{enumerate*}
A solution to the first question would yield a practical tool for
copy editors and programmers alike, automatically identifying the most
common changes.
By leveraging tools from program synthesis, such groups of edits
could be turned into interpretable rules and scripts~\citep{rolim2017learning}.
When there is no simple hard rule explaining how to apply an edit,
an answer to the second question would be of great use, \eg, to
automatically rewrite natural language following some stylistic rule.

We propose to handle edit data in an autoencoder-style framework, in
which an ``edit encoder'' $\changeReprFunc$ is trained to compute a
representation of an edit \change{\elementbefore}{\elementafter}, 
and a ``neural editor'' \editorSym is trained to construct
\elementafter from the edit representation and \elementbefore.
This framework ensures that the edit representation is semantically
meaningful, and a sufficiently strong neural editor allows this
representation to not be specific to the changed element.
We experiment with various neural architectures that can learn to
represent and apply edits and hope to direct the attention of the
research community to this new and interesting data source, leading to
better datasets and stronger models.

Briefly, the contributions of our paper are:
\begin{enumerate*}[label=(\alph*)]
  \item in \rSC{sec:task}, we present a new and important machine
    learning task on learning representations of edits 
  \item we present a family of models that capture the structure 
    of edits and compute efficient representations in \rSC{sec:model} 
  \item we create a new source code edit dataset, and release 
    the data extraction code at \url{https://github.com/Microsoft/msrc-dpu-learning-to-represent-edits}
    and the data at \url{http://www.cs.cmu.edu/~pengchey/githubedits.zip}.
  \item we perform a set of experiments on the learned edit
   representations in \rSC{sect:evaluation} for natural language text and source code and
   present promising empirical evidence that our models succeed in
   capturing the semantics of edits.
\end{enumerate*}

\section{Task}
\label{sec:task}
%!TEX root = ../editrepresentations.tex

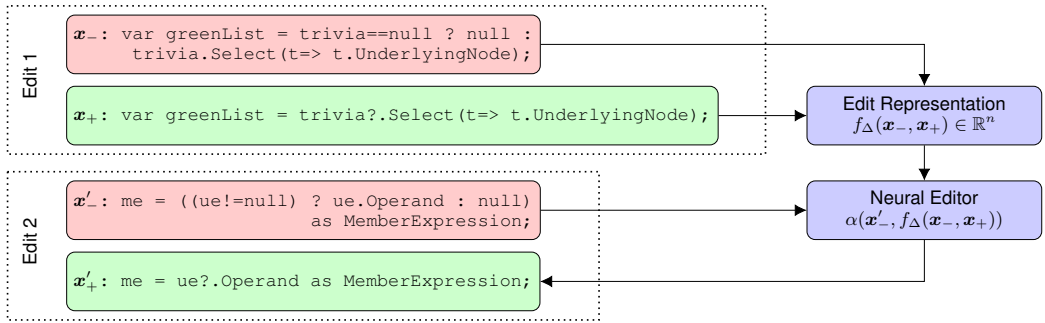
\begin{figure}\centering
\resizebox{\columnwidth}{!}{
    \begin{tikzpicture}[node distance=1.5cm,
                    >={Latex[width=2mm,length=2mm]},
                    % Specifications for style of nodes:
                    base/.style = {rectangle, rounded corners, draw=black,
                                  minimum width=4cm, minimum height=1cm,
                                  text centered, font=\footnotesize\sffamily},
                    blue/.style = {base, fill=blue!20},
                    red/.style = {base, fill=red!20, font=\footnotesize\ttfamily},
                    green/.style = {base, fill=green!20, font=\footnotesize\ttfamily},
                    every node/.style={fill=white, font=\footnotesize\sffamily},
                    align=center]
  % Specification of nodes (position, etc.)
  \node[red] (elementbefore1) at (0,0) 
   {\elementbefore: var greenList = trivia==null ? null : 
                \\\hfill trivia.Select(t=> t.UnderlyingNode);};
  \node[green, anchor=north west] (elementafter1) at ($(elementbefore1.south west) + (0, -.2)$)
   {\elementafter: var greenList = trivia?.Select(t=> t.UnderlyingNode);};
  \node (edit1) [left = of elementbefore1, xshift=25, rotate=90] {Edit 1};
  \draw[thick,dotted] ($(elementbefore1.north west)+(-1,0.15)$) rectangle ($(elementafter1.south east)+(.8,-0.15)$);

  \node[red, anchor=north west] (elementbefore2) at ($(elementafter1.south west) + (0, -.6)$)
   {$\elementbefore'$: me = ((ue!=null) ? ue.Operand : null)
                               \\\hfill as MemberExpression;};
  \node[green, anchor=north west] (elementafter2) at ($(elementbefore2.south west) + (0, -.2)$)
   {$\elementafter'$: me = ue?.Operand as MemberExpression;};
  \node (edit2) [left = of elementbefore2, xshift=25, rotate=90] {Edit 2};
  \draw[thick,dotted] ($(elementbefore2.north west)+(-1,0.15)$) rectangle ($(elementafter2.south east)+(1,-0.15)$);

  \node[blue, anchor=north west] (editrepfun) at ($(elementafter1.north east) + (1.5, 0)$)
   {Edit Representation\\$\changeReprFunc(\elementbefore, \elementafter)\in\mathbb{R}^n$};
  
  \node[blue, anchor=north west] (neuraleditor) at ($(editrepfun.south west) + (0, -.6)$)
   {Neural Editor\\$\editorFunc{\elementbefore'}{\editrepr{\elementbefore}{\elementafter}}$};

  \draw[->]     (elementbefore1) -| (editrepfun);
  \draw[->]     (elementafter1) -- (editrepfun);

  \draw[->]     (elementbefore2) -- (neuraleditor);
  \draw[->]     (editrepfun) -- (neuraleditor);

  \draw[->]     (neuraleditor) |- (elementafter2.east);
\end{tikzpicture}
}
\vspace{-3ex}
\caption{Given an edit (Edit 1) of \elementbefore to \elementafter, 
    \changeReprFunc computes an edit representation vector.
    Using that representation vector the neural editor \editorSym applies
    the same edit to a new $\elementbefore'$. The code snippets shown here
    are real code change examples from the \code{roslyn} open-source compiler project.}\label{fig:task}
\vspace{-2ex}
\end{figure}

In this work, we are interested in learning to represent and apply edits
on discrete sequential or structured data, such as text or source code 
parse trees\footnote{Existing editing systems, \eg the grammar checker in text editors and code refactoring module in IDEs, are powered by domain-specific, manually crafted rules, while we aim for a data-driven, domain-agnostic approach.}. \autoref{fig:task} gives a graphical overview of the task,
described precisely below.

\paragraph{Edit Representation}
Given a dataset of edits 
 $\{\change{\elementbefore^{(i)}}{\elementafter^{(i)}}\}_{i=1}^N,$
where $\elementbefore^{(i)}$ is the original version of some object and
$\elementafter^{(i)}$ its edited form (see upper half of \autoref{fig:task}
for an example), our goal is to learn a representation function \changeReprFunc
that maps an edit operation \change{\elementbefore}{\elementafter}\ to a 
real-valued \emph{edit representation} 
 $\editrepr{\elementbefore}{\elementafter} \in \mathbb{R}^n.$
A desired quality of \changeReprFunc is for the computed edit representations to
have the property that semantically similar edits have nearby representations in
$\mathbb{R}^n$. 
Having distributed representations also allows other interesting downstream tasks, \eg, unsupervised clustering and visualization of similar edits from large-scale data (\eg the GitHub commit stream), which would be useful for developing human-assistance toolkits for discovering and extracting emerging edit patterns (\eg new bug fixes or emerging ``best practices'' of coding).

\paragraph{Neural Editor}
Given an edit representation function \changeReprFunc, we want to learn
to apply edits in a new context.
This can be achieved by learning a \emph{neural editor} \editorSym that
accepts an edit representation $\editrepr{\elementbefore}{\elementafter}$
and a new input $\elementbefore'$ and generates $\elementafter'$.\footnote{We leave the problem of identifying which edit representation \editrepr{\elementbefore}{\elementafter} to apply to $\elementbefore'$ as interesting future work.}
This is illustrated in the lower half of \autoref{fig:task}.

\section{Model}
\label{sec:model}
%!TEX root = ../editrepresentations.tex

We cast the edit representation problem as an autoencoding task,
where we aim to minimize the reconstruction error of \editorSym for 
the edited version \elementafter given the edit representation
\editrepr{\elementbefore}{\elementafter} and the original version
\elementbefore.
By limiting the capacity of \changeReprFunc's output and allowing the
model to freely use information about \elementbefore, we are introducing
a ``bottleneck'' that forces the overall framework to not simply treat
\editrepr{\elementbefore}{\elementafter} as an encoder of \elementafter.
The main difference from traditional autoencoders is that in our setup,
an optimal solution requires to re-use as much information as possible
from \elementbefore to make the most of the capacity of 
\changeReprFunc.
Formally, given a probabilistic editor function $P_\editorSym$ such as
a neural network and a dataset
 $\{\change{\elementbefore^{(i)}}{\elementafter^{(i)}}\}_{i=1}^N,$
we seek to minimize the negative likelihood loss
\begin{align*}
 \mathcal{L} = - \frac{1}{N}\sum_i \log P_\editorSym(\elementafter \mid \elementbefore, \changeReprFunc(\elementbefore, \elementafter)).
\end{align*}
Note that this loss function can be interpreted in two ways: 
 (1) as a conditional autoencoder that encodes the salient information
     of an edit, given \elementbefore
and
 (2) as an encoder-decoder model that encodes \elementbefore and decodes
     \elementafter conditioned on the edit representation 
     \editrepr{\elementbefore}{\elementafter}.
In the rest of this section, we discuss our methods to model 
$P_\editorSym$ and \changeReprFunc as neural networks.

\subsection{Neural Editor}
As discussed above, \editorSym should use as much information as
possible from \elementbefore, and hence, an encoder-decoder architecture
with the ability to copy from the input is most appropriate.
As we are primarily interested in edits on text and source code in this
work, we explored two architectures: a sequence-to-sequence model for
text, and a graph-to-tree model for source code, whose known semantics
we can leverage both on the encoder as well as on the decoder side.
Other classes of edits, for example, image manipulation, would most
likely be better served by convolutional neural models.

\paragraph{Sequence-to-Sequence Neural Editor}
First, we consider a standard sequence-to-sequence model with attention
(over the tokens of \elementbefore).
The architecture of our sequence-to-sequence model is similar to that of
\citet{luong2015effective}, with the difference that we use a bidirectional
LSTM in the encoder and a token-level copying mechanism~\citep{vinyals2015pointer}
that directly copies tokens into the decoded sequence.
Whereas in standard sequence-to-sequence models the decoder is initialized
with the representation computed by the encoder, we initialize it with the
concatenation of encoder output and the edit representation.
We also feed the edit representation as input to the decoder LSTM at each 
decoding time step.
This allows the LSTM decoder to take the edit representation into consideration
while generating the output sequence.

\paragraph{Graph-to-Tree Neural Editor}
Our second model aims to take advantage of the additional structure of
\elementbefore and \elementafter.
To achieve this, we combine a graph-based encoder with a tree-based decoder.
We use $T(\element)$ to denote a tree representation of an element, \eg, the
abstract syntax tree (AST) of a fragment of source code.
We extend $T(\element)$ into a graph form $G(\element)$ by encoding additional
relationships (\eg, the ``next token'' relationship between terminal nodes, \etc) 
(see \autoref{fig:GenerationGraph}(a)).
To encode the elements of $G(\elementbefore)$ into vector representations, we use
a gated graph neural network (GGNN) \citep{li2015gated}.
Similarly to recurrent neural networks for sequences (such as biRNNs), GGNNs compute a
representation for each node in the graph, which can be used in the attention
mechanisms of a decoder.
Additionally, we use them to obtain a representation of the full input
\elementbefore, by computing their weighted average following the strategy
of \citet{gilmer2017neural} (\ie, computing a score for each node,
normalizing scores with a softmax, and using the resulting values
as weights).

\pgfmathsetmacro{\nodeToLabelSep}{0.1}
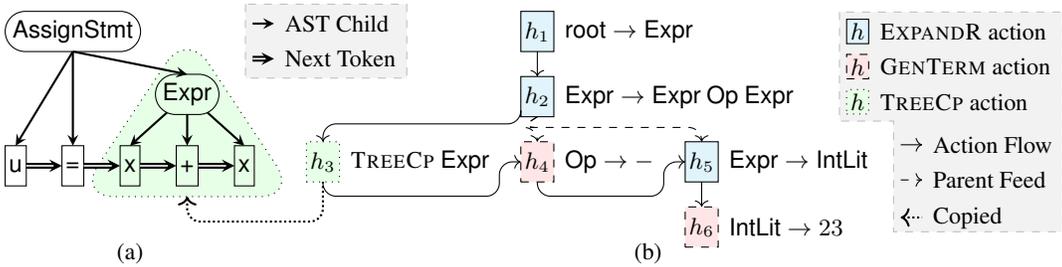
\begin{figure}
    \center
    \begin{tikzpicture}[diff, scale=0.95]
      %%%%%%%%%%%%%%%%%%%%%%%%%%%%%%% Graph of Statement
      \node[nontermNode] (Assign) at (0, 0)
        {AssignStmt};
      \node[termNode]    (Var1)   at ($(Assign.south) + (-\asthdist,\vdistTwoLvl)$)
        {u};
      \node[termNode]    (Eq)     at ($(Assign.south) + (0, \vdistTwoLvl)$)
        {=};
      \node[nontermNode] (Expr)   at ($(Assign.south) + (2*\asthdist, \vdistOneLvl)$)
        {Expr};
      \node[termNode]    (Var2)   at ($(Assign.south) + (\asthdist, \vdistTwoLvl)$)
        {x};
      \node[termNode]    (Plus)   at ($(Assign.south) + (2*\asthdist, \vdistTwoLvl)$)
        {+};
      \node[termNode]    (Var3)   at ($(Assign.south) + (3*\asthdist, \vdistTwoLvl)$)
        {x};
  
      \begin{pgfonlayer}{bg}
        \draw[rounded corners=5mm, dotted, fill=green!10]
          ($(Expr.north) + (0,.6)$)%
          --($(Var2.south west) + (-0.6,-0.15)$)%
          --($(Var3.south east) + (0.6,-0.15)$)%
          --cycle;
      \end{pgfonlayer}
  
      \draw[]
        (Assign.south) edge[childEdge]     (Var1.north)
        (Assign.south) edge[childEdge]     (Eq.north)
        (Assign.south) edge[childEdge]     (Expr.north)
        (Expr)         edge[childEdge]     (Var2.north)
        (Expr)         edge[childEdge]     (Plus.north)
        (Expr)         edge[childEdge]     (Var3.north)
        (Var1.east)    edge[nextTokenEdge] (Eq.west)
        (Eq.east)      edge[nextTokenEdge] (Var2.west)
        (Var2.east)    edge[nextTokenEdge] (Plus.west)
        (Plus.east)    edge[nextTokenEdge] (Var3.west)
        %(Var3.south)   edge[lastUseEdge, bend left=10] (Var2.south)
      ;
  
      %%%%%%%%%%%%%%%%%%%%%%%%%%%%%%% Expansion Of Tree
      \node[expansionState, ruleExpansionNode] (step1) at ($(Assign) + (6.5,0)$)
        {$h_1$};
      \node[expansionLabel, anchor=west] (step1-label) at ($(step1.east) + (\nodeToLabelSep,0)$)
        {$\textsf{root} \rightarrow \textsf{Expr}$};
      \node[expansionState, ruleExpansionNode, anchor=north east] (step2) at ($(step1.south east) + (0,-.35)$)
        {$h_2$};
      \node[expansionLabel, anchor=west] (step2-label) at ($(step2.east) + (\nodeToLabelSep,0)$)
        {$\textsf{Expr} \rightarrow \textsf{Expr}~\textsf{Op}~\textsf{Expr}$};

      \node[expansionState, copyActionNode, anchor=north east] (step3) at ($(step2.south east) + (-3,-.35)$)
        {$h_3$};
      \node[expansionLabel, anchor=west] (step3-label) at ($(step3.east) + (\nodeToLabelSep,0)$)
        {\subtreecopy \textsf{Expr}};
      
      \node[expansionState, litGenNode, anchor=north east] (step4) at ($(step2.south east) + (0,-.35)$)
        {$h_4$};
      \node[expansionLabel, anchor=west] (step3-label) at ($(step4.east) + (\nodeToLabelSep,0)$)
        {$\textsf{Op} \rightarrow -$};

      \node[expansionState, ruleExpansionNode, anchor=north east] (step5) at ($(step2.south east) + (2.3,-.35)$)
        {$h_5$};
      \node[expansionLabel, anchor=west] (step5-label) at ($(step5.east) + (\nodeToLabelSep,0)$)
        {$\textsf{Expr} \rightarrow \textsf{IntLit}$};
      \node[expansionState, litGenNode, anchor=north east] (step6) at ($(step5.south east) + (0,-.35)$)
        {$h_6$};
      \node[expansionLabel, anchor=west] (step4-label) at ($(step6.east) + (\nodeToLabelSep,0)$)
        {$\textsf{IntLit} \rightarrow 23$};
      \draw[expansionFlowEdge]
        (step1.south)
        -- (step2.north);
      \draw[expansionFlowEdge]
        (step2.south west)
        -- ($(step2.south west) + (0,-.1)$)
        -| (step3.north);
      \draw[copyTreeEdge]
        ($(Plus.south) + (0, -.15)$)
        -- ($(Plus.south) + (0, -.5)$)
        -| (step3.south);
      \draw[expansionFlowEdge] 
        (step3.south)
        -- ($(step3.south) + (0, -.2)$)
        -| ($(step4.west)  + (-.3, 0)$)
        -- ($(step4.west)$);
      \draw[parentFeedingEdge]
        (step2.south)
        -- ($(step2.south west) + (0,-.15)$)
        -| (step4.north);
      \draw[expansionFlowEdge] 
        (step4.south)
        -- ($(step4.south) + (0, -.2)$)
        -| ($(step5.west)  + (-.3, 0)$)
        -- ($(step5.west)$);
      \draw[parentFeedingEdge]
        (step2.south)
        -- ($(step2.south west) + (0,-.15)$)
        -| (step5.north);
        \draw[expansionFlowEdge]
        (step5.south)
        -- (step6.north);
  
      %%%%%%%%%%%%%%%%%%%%%%%%%%%%%%% Labels & Legends
      \node[] (labelA) at ($(Var2.south) + (0, -.95)$) {(a)};
      \node[] (labelB) at ($(labelA) + (7.25, 0)$) {(b)};

      \node[figureLabel, anchor=north west, minimum height=0] (label-ChildEdge) at ($(Assign.north east) + (2.25, 0)$)                           {AST Child};
      \draw[] ($(label-ChildEdge.west) + (-0.3-\nodeToLabelSep,0)$) edge[childEdge] ($(label-ChildEdge.west) + (-\nodeToLabelSep, 0)$);

      \node[figureLabel, anchor=north west, minimum height=0] (label-nextTokenEdge) at ($(label-ChildEdge.south west) + (0, -0.15)$)             {Next Token};
      \draw[] ($(label-nextTokenEdge.west) + (-0.3-\nodeToLabelSep,0)$) edge[nextTokenEdge] ($(label-nextTokenEdge.west) + (-\nodeToLabelSep, 0)$);

%      \node[figureLabel, anchor=north west, minimum height=0] (label-lastUseEdge) at ($(label-nextTokenEdge.south west) + (0, -0.15)$)           {Last Use};
%      \draw[] ($(label-lastUseEdge.west) + (-0.3-\nodeToLabelSep,0)$) edge[lastUseEdge] ($(label-lastUseEdge.west) + (-\nodeToLabelSep, 0)$);
      
      \begin{pgfonlayer}{bg}
        \draw[dashed, color=black!25, fill=gray!10]
          ($(label-ChildEdge.north west) + (-0.3-\nodeToLabelSep-0.1,0.1)$)
          rectangle 
          ($(label-nextTokenEdge.south east) + (0.1,-0.1)$);
      \end{pgfonlayer}

      % Expansion tree nodes:
      \node[expansionState, ruleExpansionNode, minimum height=0] (legendRuleExpansion) 
        at ($(step1-label.east) + (2.25,0)$)
        {$h$};
      \node[figureLabel, anchor=west] (label-legendRuleExpansion) at ($(legendRuleExpansion.east) + (\nodeToLabelSep, 0)$)
        {\expandRule action};

      \node[expansionState, litGenNode, minimum height=0, anchor=north west] (legendLitGen) 
        at ($(legendRuleExpansion.south west) + (0,-0.15)$)
        {$h$};
      \node[figureLabel, anchor=west] (label-legendLitGen) at ($(legendLitGen.east) + (\nodeToLabelSep, 0)$)
        {\genTerminal action};

      \node[expansionState, copyActionNode, minimum height=0, anchor=north west] (legendCopyAction)
        at ($(legendLitGen.south west) + (0,-0.15)$)
         {$h$};
      \node[figureLabel, anchor=west] (label-legendCopyAction) at ($(legendCopyAction.east) + (\nodeToLabelSep, 0)$)
        {\subtreecopy action};

      \node[figureLabel, anchor=north west, minimum height=0] (label-expansionFlowEdge) at ($(label-legendCopyAction.south west) + (0.75, -0.15)$) 
        {Action Flow};
      \draw[] ($(label-expansionFlowEdge.west) + (-0.3-\nodeToLabelSep,0)$) edge[expansionFlowEdge] ($(label-expansionFlowEdge.west) + (-\nodeToLabelSep, 0)$);
        
      \node[figureLabel, anchor=north west, minimum height=0] (label-parentFeedingEdge) at ($(label-expansionFlowEdge.south west) + (0, -0.15)$) {Parent Feed};
        \draw[] ($(label-parentFeedingEdge.west) + (-0.3-\nodeToLabelSep,0)$) edge[parentFeedingEdge] ($(label-parentFeedingEdge.west) + (-\nodeToLabelSep, 0)$);
  
      \node[figureLabel, anchor=north west, minimum height=0] (label-copyTreeEdge) at ($(label-parentFeedingEdge.south west) + (0, -0.15)$)      {Copied};
        \draw[] ($(label-copyTreeEdge.west) + (-0.3-\nodeToLabelSep,0)$) edge[copyTreeEdge] ($(label-copyTreeEdge.west) + (-\nodeToLabelSep, 0)$);        

      \begin{pgfonlayer}{bg}
        \draw[dashed, color=black!25, fill=gray!10]
            ($(legendRuleExpansion.north west) + (-0.1,0.1)$)
            --
            ($(legendCopyAction.south west) + (-0.1,-0.1)$)
            -|
            ($(label-copyTreeEdge.west) + (-0.4-\nodeToLabelSep,-0.2)$)
            --
            ($(label-copyTreeEdge.east) + (.8,-0.2)$)
            |-
            ($(legendRuleExpansion.north west) + (-0.1,0.1)$);
        \end{pgfonlayer}
    \end{tikzpicture}
    \vspace{-.7cm}
    \caption{\label{fig:GenerationGraph}%
     (a) Graph representation of statement \code{u = x + x}.
       Rectangular (resp. rounded) nodes denote tokens (resp. non-terminals).
     (b) Sequence of tree decoding steps yielding \code{x + x - 23},
       where \code{x + x} is copied (using the \subtreecopy action) from the context graph in (a).}
\end{figure}

%Fig 2b never referenced in text?

Our tree decoder follows the semantic parsing model of \citet{yin18emnlpdemo},
which sequentially generate a tree $T(\elementafter)$ as a series of
expansion actions $a_1 \ldots a_N$.
The probability of taking an action is modeled as $p(a_t \mid a_{<t}, s)$,
where $s$ is the input (a sequence of words in the original semantic parsing
setting) and $a_{<t}$ is the partial tree that has been generated so far.
The model of \citet{yin18emnlpdemo} mainly uses two types of actions:
\expandRule expands the current non-terminal using a grammar rule,
and \genTerminal generates a terminal token from a vocabulary or
copies a token from $s$\footnote{\expandRule corresponds to the \applyConstr action in the original model of \citet{yin18emnlpdemo}. There is also a \reduce action which marks the end of expanding a non-terminal with non-deterministic  number of child nodes. See \citet{yin18emnlpdemo} for details.}.
The dependence on the partial tree $a_{<t}$ is modeled by an LSTM cell
which is used to maintain state throughout the generation procedure.
Additionally, the LSTM receives the decoder state used to pick the
action at the parent node as an additional input (``parent-feeding'').
This process illustrated in \autoref{fig:GenerationGraph}(b).
%% MB Don't think this is crucial:
%Additionally, the model uses an attention mechanism that allows the decoder to
%attend over the input sequence $s$, similar to \citet{bahdanau2014neural}.

We extend this model to our setting by replacing the input sequence $s$
by \elementbefore; concretely, we condition the decoder on the graph-level
representation computed for $G(\elementbefore)$. %, and use the node
%representations from that graph as memories for the attention mechanism.
Additionally, we use the change representation $\changeReprFunc(\cdot)$
as an additional input to the LSTM initial state and at every decoding step.
Based on the observation that edits to source code often manipulate the
syntax tree by moving expressions around (\eg by nesting statements in
a conditional, or renaming a function while keeping its arguments), we
extend the decoding model of \citet{yin18emnlpdemo} by adding a
facility to copy entire subtrees from the input.
For this, we add a decoder action \subtreecopy.
This action is similar to standard copying mechanism known from pointer
networks~\citep{vinyals2015pointer}, but instead of copying only a single
token, it copies the whole subtree pointed to.

However, adding the \subtreecopy action means that there are
many correct generation sequences for a target tree.
This problem appears in token-copying as well, but can be easily
circumvented by marginalizing over all correct choices at each
generation step (by normalizing the probability distribution over
allowed actions to sum up those that have the same effect).
In the subtree-copying setting, the lengths of action sequences representing different choices may differ.
In our implementation we handle this problem during training by 
simply picking the generation sequence that greedily selecting \subtreecopy
actions.
%%%% I tried the following approach but found it achieved worse results.
%We follow \citet{liu2018constrained} to handle this problem during
%training and simply pick one correct generation sequence (the one
%greedily selecting \subtreecopy) but change the objective
%such that no correct decoder action choice is penalized; achieved
%by a ``many-hot'' encoding of correct choices.
%We resolve the issue by using beam search at test time and merging
%hypotheses with identical results.
 
\subsection{Edit Representation}
\label{sec:edit_encoder}

To compute a useful edit representation, a model needs to focus on the
differences between \elementbefore and \elementafter.
A risk in our framework is that \changeReprFunc degenerates into an
encoder for \elementafter, turning \editorSym into a decoder.
To avoid this, we need to follow the standard autoencoder trick,
\ie it is important to limit the capacity of the result
of \changeReprFunc by generating the edit representation in a
low-dimensional space $\mathbb{R}^N$. This acts as a bottleneck and encodes
only the information that is needed to reconstruct \elementafter from
\elementbefore.
We again experimented with both sequence-based and graph-based
representations of edits.

\begin{figure}
  \center
  \resizebox{\columnwidth}{!}{
  \begin{tikzpicture}[diff]
    \node[beforeVer, nontermNode] (1-Assign) at (0, 0) 
      {AssignStmt \nodepart{two} $\emptyset$};
    \node[beforeVer, nontermNode] (1-FieldA) at ($(1-Assign.south) + (-2*\asthdist,\vdistOneLvl)$)
      {FieldAccess \nodepart{two} $\emptyset$};
    \node[beforeVer, termNode]    (1-Var1)   at ($(1-Assign.south) + (-3*\asthdist,\vdistTwoLvl)$)
      {v\phantom{F}\!\!\! \nodepart{two} $\leftrightarrow$};
    \node[beforeVer, termNode]    (1-Dot)    at ($(1-Assign.south) + (-2*\asthdist,\vdistTwoLvl)$)
      {.\phantom{F}\!\!\! \nodepart{two} $-$};
    \node[beforeVer, termNode]    (1-FieldN) at ($(1-Assign.south) + (-1*\asthdist,\vdistTwoLvl)$)
      {F \nodepart{two} $-$};
    \node[beforeVer, termNode]    (1-Eq)     at ($(1-Assign.south) + (0, \vdistTwoLvl)$)
      {= \nodepart{two} $=$};
    \node[beforeVer, nontermNode] (1-Expr)   at ($(1-Assign.south) + (2*\asthdist, \vdistOneLvl)$)
      {Expr \nodepart{two} $=$};
    \node[beforeVer, termNode]    (1-Var2)   at ($(1-Assign.south) + (1*\asthdist, \vdistTwoLvl)$)
      {x \nodepart{two} $=$};
    \node[beforeVer, termNode]    (1-Plus)   at ($(1-Assign.south) + (2*\asthdist, \vdistTwoLvl)$)
      {+ \nodepart{two} $=$};
    \node[beforeVer, termNode]    (1-Var3)   at ($(1-Assign.south) + (3*\asthdist, \vdistTwoLvl)$)
      {x \nodepart{two} $=$};

    \node[afterVer, nontermNode]  (2-Assign) at ($(1-Assign) + (4.75, 0)$)
      {AssignStmt \nodepart{two} $\emptyset$};
    \node[afterVer, termNode]     (2-Var1)   at ($(2-Assign.south) + (-1*\asthdist,\vdistTwoLvl)$)
      {u \nodepart{two} $\leftrightarrow$};
    \node[afterVer, termNode]     (2-Eq)     at ($(2-Assign.south) + (0, \vdistTwoLvl)$)
      {= \nodepart{two} $=$};
    \node[afterVer, nontermNode]  (2-Expr)   at ($(2-Assign.south) + (2*\asthdist, \vdistOneLvl)$)
      {Expr \nodepart{two} $=$};
    \node[afterVer, termNode]     (2-Var2)   at ($(2-Assign.south) + (1*\asthdist, \vdistTwoLvl)$)
      {x \nodepart{two} $=$};
    \node[afterVer, termNode]     (2-Plus)   at ($(2-Assign.south) + (2*\asthdist, \vdistTwoLvl)$)
      {+ \nodepart{two} $=$};
    \node[afterVer, termNode]     (2-Var3)   at ($(2-Assign.south) + (3*\asthdist, \vdistTwoLvl)$)
      {x \nodepart{two} $=$};      

    \draw[]
      (1-Assign.south) edge[childEdge] (1-FieldA.north)
      (1-Assign.south) edge[childEdge] (1-Eq.north)
      (1-Assign.south) edge[childEdge] (1-Expr.north)
      (1-FieldA.south) edge[childEdge] (1-Var1.north)
      (1-FieldA.south) edge[childEdge] (1-Dot.north)
      (1-FieldA.south) edge[childEdge] (1-FieldN.north)
      (1-Expr)         edge[childEdge] (1-Var2.north)
      (1-Expr)         edge[childEdge] (1-Plus.north)
      (1-Expr)         edge[childEdge] (1-Var3.north)
      (1-Var1.east)    edge[nextTokenEdge] (1-Dot.west)
      (1-Dot.east)     edge[nextTokenEdge] (1-FieldN.west)
      (1-FieldN.east)  edge[nextTokenEdge] (1-Eq.west)
      (1-Eq.east)      edge[nextTokenEdge] (1-Var2.west)
      (1-Var2.east)    edge[nextTokenEdge] (1-Plus.west)
      (1-Plus.east)    edge[nextTokenEdge] (1-Var3.west)
      (2-Assign.south) edge[childEdge] (2-Var1.north)
      (2-Assign.south) edge[childEdge] (2-Eq.north)
      (2-Assign.south) edge[childEdge] (2-Expr.north)
      (2-Expr)         edge[childEdge] (2-Var2.north)
      (2-Expr)         edge[childEdge] (2-Plus.north)
      (2-Expr)         edge[childEdge] (2-Var3.north)
      (2-Var1.east)    edge[nextTokenEdge] (2-Eq.west)
      (2-Eq.east)      edge[nextTokenEdge] (2-Var2.west)
      (2-Var2.east)    edge[nextTokenEdge] (2-Plus.west)
      (2-Plus.east)    edge[nextTokenEdge] (2-Var3.west)
      (1-Var1.south)   edge[bend right=13, replacementEdge] (2-Var1.south)
      (1-Eq.south)     edge[bend right=7, sameEdge] (2-Eq.south)
      (1-Expr.south)   edge[bend right=7, sameEdge] (2-Expr.south)
      (1-Var2.south)   edge[bend right=7, sameEdge] (2-Var2.south)
      (1-Plus.south)   edge[bend right=7, sameEdge] (2-Plus.south)
      (1-Var3.south)   edge[bend right=7, sameEdge] (2-Var3.south)
    ;

    \def\mystrut{\phantom{Fp}\!\!\!\!\!}
    \node[diffSequenceNode, anchor=south west] (DiffSeq-1) at ($(1-Var1.south west) + (-5, 0)$)
      {                \mystrut$\leftrightarrow$
       \nodepart{two}  \mystrut\code{v}
       \nodepart{three}\mystrut\code{u}};
    \node[diffSequenceNode] (DiffSeq-2) at ($(DiffSeq-1.north east) + (0.1, 0)$)
       {                \mystrut$-$
        \nodepart{two}  \mystrut\code{.}
        \nodepart{three}\mystrut$\emptyset$};
    \node[diffSequenceNode] (DiffSeq-3) at ($(DiffSeq-2.north east) + (0.1, 0)$)
      {                \mystrut$-$
        \nodepart{two}  \mystrut\code{F}
        \nodepart{three}\mystrut$\emptyset$};
    \node[diffSequenceNode] (DiffSeq-4) at ($(DiffSeq-3.north east) + (0.1, 0)$)
      {                \mystrut$=$
       \nodepart{two}  \mystrut\code{=}
       \nodepart{three}\mystrut\code{=}};
    \node[diffSequenceNode] (DiffSeq-5) at ($(DiffSeq-4.north east) + (0.1, 0)$)
      {                \mystrut$=$
       \nodepart{two}  \mystrut\code{x}
       \nodepart{three}\mystrut\code{x}};
    \node[diffSequenceNode] (DiffSeq-6) at ($(DiffSeq-5.north east) + (0.1, 0)$)
      {                \mystrut$=$
      \nodepart{two}  \mystrut\code{+}
      \nodepart{three}\mystrut\code{+}};
    \node[diffSequenceNode] (DiffSeq-7) at ($(DiffSeq-6.north east) + (0.1, 0)$)
      {                \mystrut$=$
       \nodepart{two}  \mystrut\code{x}
       \nodepart{three}\mystrut\code{x}};
    
    \node[anchor=north] (labelA) at ($(DiffSeq-4.south) + (0, -.5)$) {(a)};
    \node[] (labelB) at ($(labelA) + (8, 0)$) {(b)};
  \end{tikzpicture}
  }
  \vspace{-.7cm}
  \caption{\label{fig:ChangeGraph}
   Sequence (a) and graph (b) representation of edit of \code{v.F = x + x}
   to \code{u = x + x}.}
\end{figure}
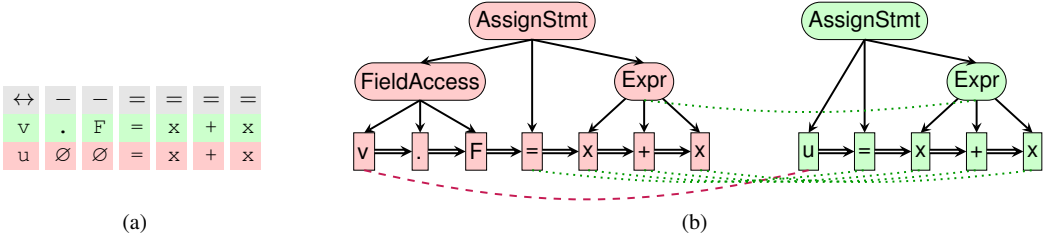

\paragraph{Sequence Encoding of Edits}
Given \elementbefore (resp.~\elementafter) as sequence of tokens
 $t_{-}^{(0)}, \ldots t_{-}^{(T_{-})}$
 (resp. $t_{+}^{(0)}, \ldots t_{+}^{(T_{+})}$),
we can use a standard (deterministic) diffing algorithm to compute an alignment of tokens 
in the two sequences. We then use extra symbols 
 $\emptyset$ for padding,
 $+$ for additions,
 $-$ for deletions,
 $\leftrightarrow$ for replacements, and
 $=$ for unchanged tokens
to generate a single sequence representing both \elementbefore and
\elementafter.
This is illustrated in \autoref{fig:ChangeGraph}(a).
By embedding the three entries in each element of the sequence separately
and concatenating their representation, they can be fed into a standard
sequence encoder whose final state is our desired edit representation.
In this work, we use a biLSTM.

% ``='' missing in Fig3a

\paragraph{Graph Encoding of Edits}
As in the graph-to-tree neural editor, we represent \elementbefore and
\elementafter as trees $T(\elementbefore)$ and $T(\elementafter)$.
We combine these trees into a graph representation
$G(\change{\elementbefore}{\elementafter})$ by merging both trees into
one graph, using ``Removed'', ``Added'' and ``Replaced'' edges.
To connect the two trees, we compute the same alignment as in the sequence case,
connecting leaves that are the same and each replaced leaf to its replacement.
We also propagate this information up in the trees, \ie, two inner nodes are
connected by ``='' edges if all their descendants are connected by ``='' edges.
This is illustrated in \autoref{fig:ChangeGraph}(b).
Finally, we also use the same ``+'' / ``-'' / ``$\leftrightarrow$'' / ``='' tags
for the initial node representation, computing it as the concatenation of the
string label (\ie token or non-terminal name) and the embedding of the tag.
To obtain an edit representation, we use a GGNN unrolled for a fixed number
of timesteps and again use the weighted averaging strategy of
\citet{gilmer2017neural}.

\section{Evaluation}
\label{sect:evaluation}
%!TEX root = ../editrepresentations.tex

Evaluating an unsupervised representation learning method
is challenging, especially for a newly defined task. 
Here, we aim to evaluate the quality of the learned edit
representations with a series of qualitative and quantitative
metrics on natural language and source code.

\subsection{Datasets and Configuration}
\paragraph{Natural Language Edits}
We use the \textbf{WikiAtomicEdits}~\citep{faruqui2018wikiatomicedits} dataset
of pairs of short edits on Wikipedia articles.
We sampled 1040$K$ edits from the English \textit{insertion} portion of the dataset
and split the samples into $1000K$/$20K$/$20K$ train-valid-test sets.

\paragraph{Source Code Edits}
To obtain a dataset for source code, we clone a set of 54 C\# projects on GitHub and collected 
a \textbf{GitHubEdits} dataset (see \autoref{app:datasets} for more information).
We selected all changes in the projects that are no more than 3 lines long and
whose surrounding 3 lines of code before and after the edited lines have \emph{not} been 
changed, ensuring that the edits are separate and short.
We then parsed the two versions of the source code and take as \elementbefore and
\elementafter the code that belongs to the top-most AST node that contains the
edited lines.
Finally, we remove trivial changes such variable renaming, changes within comments or formatting changes.
Overall, this yields 111\,724 edit samples.
For each edit we run a simple C\# % semantic [MB] I'm uncomfortable calling the lexer a semantic analysis
analysis to detect all variables and normalize variable names such that each
unique variable within \elementbefore and \elementafter has a unique normalized 
name \code{V0}, \code{V1}, \etc
This step is necessary to avoid the sparsity of data induced by the variety of
different identifier naming schemes.
We split the dataset into 91,372~/~10,176~/~10,176 samples as train/valid/test sets.

Additionally, we introduce a labeled dataset of source code edits by using
C\# ``fixers''.
Fixers are small tools built on top of the C\# compiler, used to
perform common refactoring and modernization tasks (\eg, using new syntactic
sugar).
We selected 16 of these fixers and ran them on 6 C\#
projects to generate a small \textbf{C\#Fixers} dataset of 2,878 edit pairs with known semantics.
We present descriptions and examples of each fixer in~\autoref{app:datasets}.

%\paragraph{Encoding Context Information} 
\paragraph{Configuration}
Throughout the evaluation we use a fixed size of 512 for edit representations.
The size of word embeddings and hidden states of encoding LSTMs is 128.
The dimensionality of the decoding LSTM is set to 256.
Details of model configuration can be found in~\rSC{app:datasets}.

When generating the target \elementafter, our neural editor model can 
optionally take as input the context of the original input \elementbefore (e.g., 
the preceding and succeeding code segments surrounding \elementbefore), whose information could be useful for predicting \elementafter.
For example, in source code edits the updated code snippet \elementafter may reuse variables defined in the preceding snippet.
In our code experiments, we use a standard bidirectional LSTM network to encode the tokenized 3 lines of code before and after \elementbefore as context. The encoded context is used to initialize the decoder, and as an additional source for the pointer network to copy tokens from.

\subsection{Quality of Edit Representations}
\label{sec:exp:edit_repr_quality}

First, we study the ability of our models to encode edits in a semantically
meaningful way.

\begin{wrapfigure}[20]{r}{.48\textwidth}
    \vspace{-2ex}
    \includegraphics[width=\linewidth]{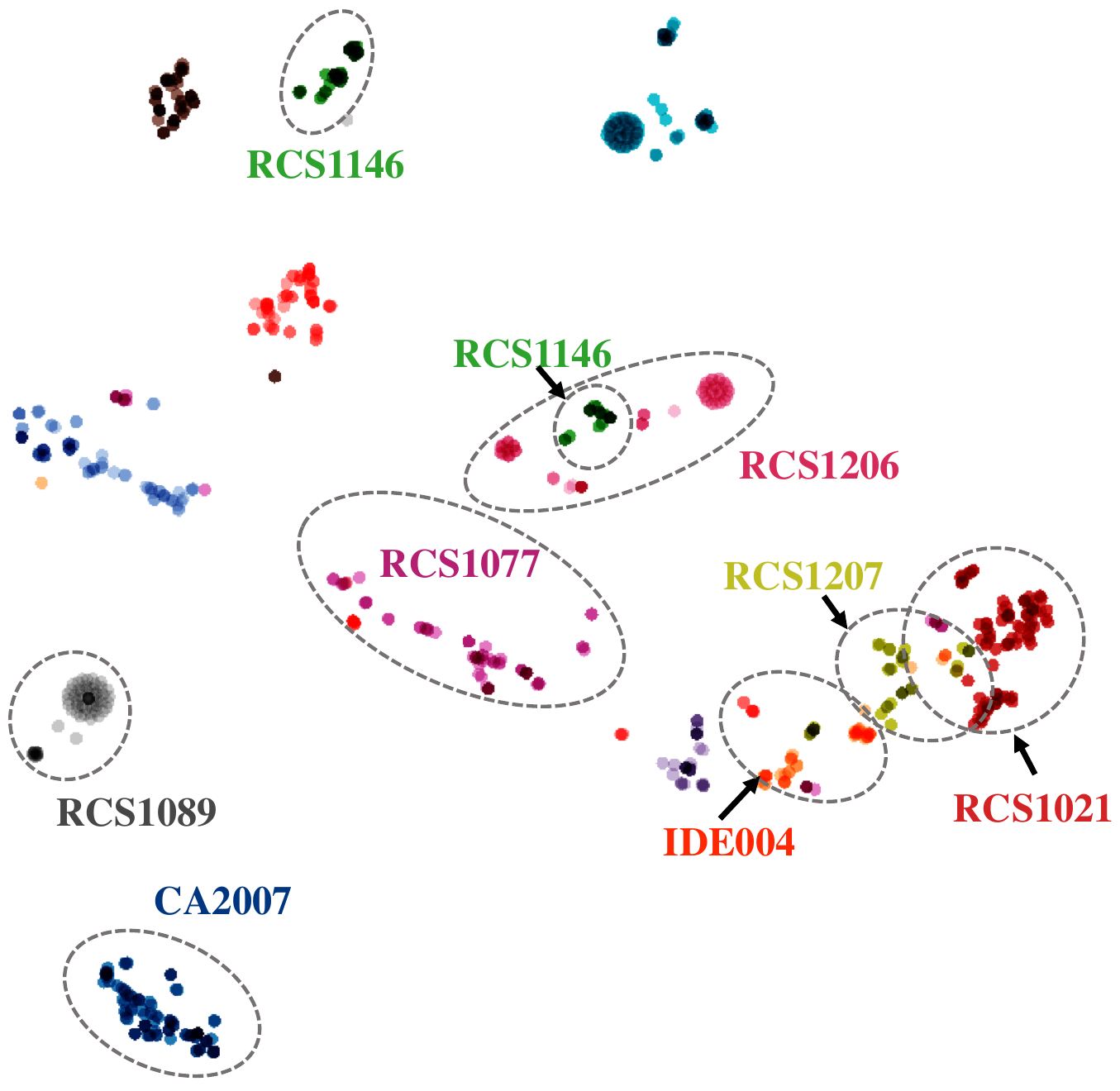}
    \vspace{-4ex}
    \caption{\label{fig:edit-tsne}%
     t-SNE visualization of edits from 13 C\# fixers, where point color
     indicates the fixer. Labels indicate the id of the fixer, see main text.}
\end{wrapfigure}
\paragraph{Visualizing Edits on Fixers Data} In a first experiment, 
we train our sequential neural editor model on our GitHubEdits data and then
compute representations for the edits generated by the C\# fixers.
A t-SNE visualization~\citep{maaten2008visualizing} of the encodings is shown in \autoref{fig:edit-tsne}.
For this visualization, we randomly selected 100 examples from the edits
of each fixer (if that fixer has more than 100 samples) and discarded fixer categories with less than 40 examples.
Readers are referred to~\autoref{app:datasets} for detailed descriptions of each fixer category.
We find that our model produces dense clusters for simple or distinctive code edits, 
 \eg fixer \href{https://github.com/JosefPihrt/Roslynator/blob/master/docs/analyzers/RCS1089.md}{\texttt{RCS1089}}
  (using the \code{++} or \code{--} unary operators instead of a binary operator (\eg, \change{\code{i = i + 1}}{\code{i++}}),
 and fixer \texttt{CA2007} (adding \code{.ConfigureAwait(false)} for \code{await} statements).
We also analyzed cases where (1) the edit examples from the same fixer are scattered, 
or (2) the clusters of different fixers overlap with each other.
For example, the 
 fixer \href{https://github.com/JosefPihrt/Roslynator/blob/master/docs/analyzers/RCS1077.md}{\texttt{RCS1077}}
covers 12 different aspects of optimizing LINQ method calls (\eg, type casting, counting, etc.),
and hence its edits are scattered.
On the other hand,
 fixers \href{https://github.com/JosefPihrt/Roslynator/blob/master/docs/analyzers/RCS1146.md}{\texttt{RCS1146}}
 and \href{https://github.com/JosefPihrt/Roslynator/blob/master/docs/analyzers/RCS1206.md}{\texttt{RCS1206}} 
yield overlapping clusters, as both fixers change code to use the \code{?.}
operator.
Fixers
 \href{https://github.com/JosefPihrt/Roslynator/blob/master/docs/analyzers/RCS1207.md}{\texttt{RCS1207}} 
 (change a lambda to a method group, \eg \change{\code{foo(x=>bar(x))}}{\code{foo(bar)}})
and
 \href{https://github.com/JosefPihrt/Roslynator/blob/master/docs/analyzers/RCS1021.md}{\texttt{RCS1021}}
 (simplify lambda expressions, \eg \change{\code{foo(x=>\{return 4;\})}}{\code{foo(x=>4)}})
are similar, as both inline lambda expressions in two different ways.
%(change a lambda to a method group, \eg \code{(x) => foo(x)} $\mapsto$ \code{foo}) and
%(simplifying lambda expressions, \eg \code{x => \{return foo(x)\};} $\mapsto$ \code{x => foo(x)}), 
Analysis yields that the representation is highly dependent on surface tokens.
For instance, \texttt{IDE004} (removing redundant type casts, \eg 
 \change{\code{(int)2}}{\code{2}})
and 
 \href{https://github.com/JosefPihrt/Roslynator/blob/master/docs/analyzers/RCS1207.md}{\texttt{RCS1207}}
  (removing explicit argument lists)
yield overlapping clusters, as both involve deleting identifiers wrapped by parentheses.

\begin{table}[t] 
\footnotesize
\begin{tabular}{@{}lp{6cm}p{6cm}@{}} \toprule
% \multicolumn{2}{l}{Bag of Words Model} & Seq2Seq -- Seq Edit Encoder \\ \midrule
                    & Bag of Words Model & Seq2Seq -- Seq Edit Encoder\\ \midrule
 \multicolumn{2}{l}{DCG/NDCG$@5$ ~~~~~~~~~~~ 9.3 / 67.3\%}  & 13.5 / 90.3\% \\
 \multicolumn{2}{l}{DCG$@5$ (by edit size) ~~ 1: 14.7~~~2-3: 10.8~~~$>$3: 5.4} & 1: 16.2~~~2-3: 12.9~~~$>$3:  12.4\\
% \scriptsize NDCG@5  & 67.3\%    & 90.3\% \\ 

\toprule

\multicolumn{3}{p{\columnwidth}}{\textbf{Example 1}~\insertText{daniel james nava ( born february 22 , 1983 ) is an american professional baseball outfielder} nava is only the fourth player in mlb history to hit a grand slam in his first major league at bat and the second to do it on the first pitch .} \\
NN-1 & he batted .302 with 73 steals , and received a september call - up to the major leagues \insertText{as an outfielder} . & ~\insertText{arthur ray briles ( born december 3 , 1955 ) is a former american football coach and} his most recent head coaching position was at baylor university , a position he held from the 2008 season through the 2015 season . \\
NN-2 & he played \insertText{as an outfielder} for the hanshin tigers . & ~\insertText{jonathan david disalvatore ( born march 30 , 1981 ) is a professional ice hockey} he was selected by the san jose sharks in the 4th round ( 104th overall ) of the 2000 nhl entry draft .\\
NN-3 & in 2012 , his senior at oak mountain , dahl had a .412 batting average , 34 runs batted in ( rbis ) , and 18 stolen bases \insertText{as an outfielder .} &~\insertText{professor paul talalay ( born march 31 , 1923 ) is the john jacob abel} distinguished service professor of pharmacology and director of the laboratory for molecular sciences at johns hopkins school of medicine in baltimore .\\ \toprule

\multicolumn{3}{p{\columnwidth}}{\textbf{Example 2} she , along with her follow artist carolyn mase studied with \insertText{impressionist landscape painter} john henry twachtman at the art students league of new york .}\\
NN-1 & his brother was draughtsman william daniell and his uncle was \insertText{landscape painter} thomas daniell . & the first painting was a portrait of a young girl , emerantia van beresteyn , the sister of \insertText{the landscape painter} nicolaes van beresteyn , the later founder of half of this hofje .\\
NN-2 & william james linton ( december 7 , 1812 - december 29 , 1897 ) was an english - born american wood engraver ,  \insertText{landscape painter ,} political reformer and author of memoirs , novels , poetry and non-fiction . & he was the club 's top scorer with 22 goals in all competitions , one more than \insertText{senegalese striker} lamine diarra , who left the club at the end of the season .\\
NN-3 & early on , hopper modeled his style after chase and french \insertText{impressionist} masters édouard manet and edgar degas . & caforio '' aggressively attacked '' his opponent , \insertText{republican incumbent} steve knight , for his delayed response to the leak .\\

\bottomrule
\end{tabular}
\caption{\label{fig:nl-dcg}Natural language human evaluation results and 3 nearest neighbors. 
 \insertText{Inserted text} marked. \textbf{Example 1} neural editing model returns syntactically and semantically similar edits. \textbf{Example 2} Neural edit representations are sensitive to position.}
\vspace{-2mm} 
\end{table}

\paragraph{Human Evaluation on Encoding Natural Language WikiAtomicEdits}
In a second experiment, we test how well neighborhoods in edit representation
space correspond to semantic similarity.
We computed the five nearest neighbors of 200 randomly sampled seed edits from our training set,
using both our trained sequence-to-sequence editing model with sequential edit encoder, as well as a simple bag-of-words baseline based on TF-IDF scores.
We then rated the quality of the retrieved neighbors on a scale of
0 (``unrelated edit''), 1 (``similar edit'') and 2 (``semantically or syntactically same edit'').
Details of the annotation schema is included in~\rSC{app:annotation}.
We show the (normalized) discounted cumulative gain (DCG, \citet{manning08irbook}) for the two models at the top of \rTab{fig:nl-dcg} (higher is better).
The relevance scores indicate that our neural model clearly outperforms the simplistic baseline. 
% in grouping semantically related edits.
%Some examples of the found clusters of changes are shown in \rF{fig:nl-dcg},
%which show that the neural model succeeded in learning semantics of changes.
\rTab{fig:nl-dcg} also presents two example edits with their
nearest neighbors.
%, which show that the neural model succeeded in learning 
%semantics of edits for both $n$-grams (upper example) and complex clauses (lower example).
%\paragraph{Observation}
Example 1 shows that the neural edit models succeeded in representing syntactically and semantically similar edits, while the bag-of-words baseline relies purely on surface token overlap.
Interestingly, we also observed that the edit representations learned by the neural editing model on WikiAtomicEdits are somewhat sensitive to position, \ie the position of the
inserted tokens in both the seed edit and the nearest neighbors is similar.
This is illustrated in Example 2, where the
second (\textit{``senegalese striker''}) and the third (\textit{``republican incumbent''})
nearest neighbors returned by the neural model have similar editing positions
as the seed edit, while they are semantically diverse.  
% On average, we find that the absolute distance of the editing
% positions between a seed edit and its neighbors is 2.14 and 9.31 for the neural model
% and the Bag of Words baseline, respectively. \yi{What is the unit of the distance? Is this in the representation space? is so, it doesn't tell us something; delete}

\subsection{Edit Encoder Performance}
To evaluate the performance of our two edit encoders discussed in~\rSC{sec:edit_encoder} and 
disentangle it from the choice of neural editor, we train various combinations
of our neural editor model and manually evaluate the quality of the 
edit representation.
More specifically, we trained our neural editor models on GitHubEdits and randomly sampled 200 seed edits and computed their 3
nearest neighbors using each end-to-end model.
We then rated the resulting groups using the same 0-2 scale as above. 
%a scale of 0 (``unrelated edit''), 1 (``similar edit'') and 2 (``highly related or the same edit'').
The resulting relevance scores are shown in \rTab{tab:exp:edit_encoding_code_data}.

\begin{table}[tb]
	\caption{Relevance scores of human evaluation on GitHubEdits data. Acc.@1 denotes the ratio that the 1-nearest neighbor has a score $2$.}
	\label{tab:exp:edit_encoding_code_data}
	\centering
    \footnotesize
	\begin{tabular}{lrrr}
	\toprule
	\textbf{Model} & \textbf{DCG@3} & \textbf{NDCG@3} (\%) & \textbf{Acc.@1} (\%) \\
	\midrule
		BoW & 7.77 & 75.99 & 58.46 \\
		Seq2Seq -- Seq Edit Encoder & 10.09 & 90.05 & 75.90 \\
		Graph2Tree -- Seq Edit Encoder & {\bf 10.56} & {\bf 91.40} & {\bf 79.49} \\
		Graph2Tree -- Graph Edit Encoder & 9.44 & 86.20 & 72.31 \\
	\bottomrule
    \end{tabular}
    \vspace{-3ex}
\end{table}

\begin{table}[t]
  \caption{Two example source code edits and their nearest neighbors based on
  the edit representations computed by each model.}\label{tbl:codesamples}
    \resizebox{0.5\columnwidth}{!}{
    \begin{tabular}{lp{8cm}} 
    \multicolumn{2}{c}{Example 1} \\
    \toprule
     &\elementbefore: \lstinline|V0.SendSelectSoundRequest((int)V1);| \\
     &\elementafter:  \lstinline|V0.SendSelectSoundRequest(V1);|\\ \midrule
    
    \multicolumn{2}{l}{\underline{Seq2Seq -- Seq Edit Encoder}} \\
    $\blacktriangleright$ & \elementbefore: \lstinline|V0.Debug(() => LITERAL);|\\
                          & \elementafter: \lstinline|V0.Debug(LITERAL);|\\ \rowspace
    $\blacktriangleright$ & \elementbefore: \lstinline|V0.Debug(() => LITERAL);|\\
                          & \elementafter: \lstinline|V0.Debug(LITERAL);|\\ \rowspace
    $\blacktriangleright$ & \elementbefore: \lstinline|V0.WriteCompressedInteger((uint)V1);|\\
                          & \elementafter: \lstinline|V0.WriteCompressedInteger(V1);|\\
    \midrule
    
    \multicolumn{2}{l}{\underline{Graph2Tree -- Seq Edit Encoder}} \\
    $\blacktriangleright$ & \elementbefore: \lstinline|V0.WriteCompressedInteger((uint)V1);|\\
                          & \elementafter: \lstinline|V0.WriteCompressedInteger(V1);|\\ \rowspace
    $\blacktriangleright$ & \elementbefore: \lstinline|V0.WriteCompressedInteger((uint)V1);|\\
                          & \elementafter: \lstinline|V0.WriteCompressedInteger(V1);|\\ \rowspace
    $\blacktriangleright$ & \elementbefore: \lstinline|V0.WriteCompressedInteger((uint)V1);|\\
                          & \elementafter: \lstinline|V0.WriteCompressedInteger(V1);|\\
    \midrule
    
    \multicolumn{2}{l}{\underline{Graph2Tree -- Graph Edit Encoder}} \\
    $\blacktriangleright$ & \elementbefore: \lstinline|V0.UpdateLastRead(this.V1);|\\
                          & \elementafter: \lstinline|V0.UpdateLastRead(V1);|\\ \rowspace
    $\blacktriangleright$ & \elementbefore: \lstinline|V0.UpdateLastWrite(this.V1);|\\
                          & \elementafter: \lstinline|V0.UpdateLastWrite(V1);|\\ \rowspace
    $\blacktriangleright$ & \elementbefore: \lstinline|V0.Append(this.V1);|\\
                          & \elementafter: \lstinline|V0.Append(V1);|\\
        \bottomrule
    \end{tabular}
    }% resizebox
    \resizebox{0.5\columnwidth}{!}{
    \begin{tabular}{lp{8cm}} 
    \multicolumn{2}{c}{Example 2} \\
    \toprule
     &\elementbefore: \lstinline|string V0; string V1;| \\
     &\elementafter:  \lstinline|string V1; string V0;| \\  \midrule
    
    \multicolumn{2}{l}{\underline{Seq2Seq -- Seq Edit Encoder}} \\
    $\blacktriangleright$ & \elementbefore: \lstinline|RetryConfig V0; string V1;|\\
                          & \elementafter: \lstinline|string V1; RetryConfig V0;|\\ \rowspace
    $\blacktriangleright$ & \elementbefore: \lstinline|string[] V0; string[] V1; int V2;|\\
                          & \elementafter: \lstinline|int V2; string[] V0; string[] V1;|\\ \rowspace
    $\blacktriangleright$ & \elementbefore: \lstinline|Type V0= null; BindingFlags V1= 0;|\\
                          & \elementafter: \lstinline|BindingFlags V1= 0; Type V0= null;|\\
    \midrule

    \multicolumn{2}{l}{\underline{Graph2Tree -- Seq Edit Encoder}} \\
    $\blacktriangleright$ & \elementbefore: \lstinline|RetryConfig V0; string V1;|\\
                          & \elementafter: \lstinline|string V1; RetryConfig V0;|\\ \rowspace
    $\blacktriangleright$ & \elementbefore: \lstinline|string[] V0; string[] V1; int V2;|\\
                          & \elementafter: \lstinline|int V2; string[] V0; string[] V1;|\\ \rowspace
    $\blacktriangleright$ & \elementbefore: \lstinline|int V0 = V1; int V2 = V3;|\\
                          & \elementafter: \lstinline|int V2 = V3; int V0 = V1;|\\                       
    \midrule
    
    \multicolumn{2}{l}{\underline{Graph2Tree -- Graph Edit Encoder}} \\
    $\blacktriangleright$ & \elementbefore: \lstinline|RetryConfig V0; string V1;|\\
                          & \elementafter: \lstinline|string V1; RetryConfig V0;|\\ \rowspace
    $\blacktriangleright$ & \elementbefore: \lstinline|int V0 = V1; int V2 = V3;|\\
                          & \elementafter: \lstinline|int V2 = V3; int V0 = V1;|\\ \rowspace
    $\blacktriangleright$ & \elementbefore: \lstinline|double V0= -1; double V1= -1;|\\
                          & \elementafter: \lstinline|double V1= -1; double V0= -1;|\\ 
        \bottomrule
    \end{tabular}
    } % resizebox
\end{table}

Comparing the sequential edit encoders trained with Seq2Seq and Graph2Tree editors, we found that the edit encoder trained with the Graph2Tree objective performs better.
We hypothesize that this is because the Graph2Tree editor better captures structural-level information about an edit.
For instance, Example 1 in \rTab{tbl:codesamples} removes explicit type casting.
The Seq2Seq editor has difficulty distinguishing this type of edit, confusing it
with changes of lambda expressions to method groups (1st and 2nd nearest neighbors) %(as in \texttt{RCS1207}, \rSC{sec:exp:edit_repr_quality}),
since both two types of edits involve removing paired parentheses.

Surprisingly, we found that the graph-based edit encoder does not outperform the
sequence-based encoder.
However, we observe that the graph edit encoder in many cases tends to better capture high-level and abstract
structural edit patterns.
Example 2 in \rTab{tbl:codesamples} showcases a seed edit that swaps two consecutive declarations,
which corresponds to swapping the intermediate \texttt{Expression} nodes representing each statement on the underlying AST. 
In this case, the graph edit encoder is capable of grouping semantically similar edits, while it seems to be more
difficult for the sequential encoder encoder to capture the edit pattern.
On the other hand, we found that the graph edit encoder often fails to capture simpler, lexical level edits (\eg, Example 1).
This might suggest that terminal node information is not effectively propagated,
an interesting issue worth future investigation.

\subsection{Precision of Neural Editors}
\label{sec:perf_neural_editor}

Finally, we evaluate the performance of our end-to-end system by predicting
the edited input \elementafter given \elementbefore and the edit representation.
We are interested in answering two research questions:
{\it First}, how well can our neural editors generate \elementafter given the gold-standard edit representation
 \editrepr{\elementbefore}{\elementafter}?
{\it Second}, and perhaps more interestingly, can we use the representation of a similar edit \editrepr{\elementbefore'}{\elementafter'} to generate \elementafter by \emph{applying} that edit to \elementbefore (\ie $\hat{\element}_{+} = \editorFunc{\elementbefore}{\editrepr{\elementbefore'}{\elementafter'}}$)?

\begin{table}[tb]
	\caption{Test performance of different neural editors.}
	\label{tab:exp:neural_editor_precision}
	\centering \footnotesize

	\begin{tabular}{llrrr}
	\toprule
	&\textbf{Model} & \textbf{Acc.@1} (\%) & \textbf{Recall@5} (\%) & \textbf{PPL} per token \\
    \midrule
        \multicolumn{3}{l}{\underline{GitHubEdits}} \\
      &Seq2Seq -- Bag-of-Edits Encoder & 44.05 & 54.97 & 1.4808 \\
		  &Seq2Seq -- Seq Edit Encoder& {\bf 59.63} & {\bf 65.46} & {\bf 1.2792} \\
      &Graph2Tree -- Bag-of-Edits Encoder & 40.66 & 49.42 & 1.5058 \\
		  &Graph2Tree -- Seq Edit Encoder & 57.49 & 62.94 & 1.3043 \\
      &Graph2Tree -- Graph Edit Encoder & 48.05 & 56.51 & 1.3712 \\ 
        \multicolumn{3}{l}{\underline{WikiAtomicEdits}} \\
        &Seq2Seq -- Bag-of-Edits Encoder & 23.73 & 43.47 & 1.3730 \\
        &Seq2Seq -- Seq Edit Encoder & {\bf 72.94} & {\bf 76.53} & {\bf 1.0527} \\
	\bottomrule
	\end{tabular}
\end{table}

To answer the first question, we trained our neural editor models on the WikiAtomicEdits and the GitHubEdits dataset, and evaluate the performance of encoding and applying edits on test sets.
For completeness, we also evaluated the performance of our neural editor models with a simple ``Bag-of-Edits'' edit encoding scheme, where \editrepr{\elementbefore}{\elementafter} is modeled as the concatenation of two vectors, each representing the sum of the embeddings of added and deleted tokens in the edit, respectively.
This edit encoding method is reminiscent of the model used in~\citet{guu2017generating} for solving a different task of language modeling by marginalizing over latent edits, which we will elaborate in~\rSC{sect:rel_work}.
%However, we remark that our work is still different from 
%\py{mention we are different from Guu et all paper.}
\rTab{tab:exp:neural_editor_precision} lists the evaluation results.
With our proposed sequence- and graph-based edit encoders, our neural editor models achieve reasonable end-to-end performance, surpassing systems using bag-of-edits representations.
This is because many edits are \emph{context-sensitive} and \emph{position-sensitive}, requiring edit representation models that go beyond the bag-of-edits scheme to capture those effects (more analysis is included in~\autoref{app:clustering_exp}). 
% On WikiAtomicEdits, our Seq2Seq editor with the sequential edit encoder achieves an accuracy of 72.9\%.
%demonstrating that the neural editor is capable of using the edit representation to make high-quality edits.
Interestingly, on the GitHubEdits dataset, we find that the Seq2Seq editor with sequential edit encoder registers the best performance.
However, it should be noted that in this set of experiments, we encode the gold-standard edit \editrepr{\elementbefore}{\elementafter} to predict \elementafter.
As we will show later, better performance with the gold-standard edit does \emph{not} necessarily imply better (more generalizable) edit representation.
%However, it should be noted that better performance does \emph{not} necessarily imply better
%(more generalizable) edit representation, since we encode the gold-standard edit \editrepr{\elementbefore}{\elementafter} to predict \elementafter.
Nevertheless, we hypothesize that the higher accuracy of the Seq2Seq edit is due to the fact that a significant proportion
of edits in this dataset is small and primarily syntactically simple.
Indeed we find that 69\% of test examples have a token-level edit distance of less than 5.

To answer the second question, we use the trained neural editors from the previous experiment, and test their performance in a ``one-shot'' transfer learning scenario.
Specifically, we use our high-quality C\#Fixers dataset, and for each fixer category $\mathcal{F}$ of semantically similar edits,
we randomly select a seed edit $\{\change{\elementbefore'}{\elementafter'}\} \in \mathcal{F}$,
and use its edit representation \editrepr{\elementbefore'}{\elementafter'} to predict the updated code for
\emph{all} examples in $\mathcal{F}$, \ie, we have 
$ \hat{\element}_{+} = \editorFunc{\elementbefore}{\editrepr{\elementbefore'}{\elementafter'}}, \forall~ \{\change{\elementbefore}{\elementafter}\} \in \mathcal{F}$.
This task is highly non-trivial, since a fixer category could contain more than hundreds of edit examples collected from different C\# projects.
Therefore,
it requires the edit representations to generalize and transfer well, while being invariant of local lexical information like specific method names.
To make the experimental evaluation more robust to noise, for each fixer category $\mathcal{F},$
we randomly sample 10 seed edit pairs $\{\elementbefore'\rightarrow \elementafter' \}$, compute their edit representations and
use them to predict the edited version of the examples in $\mathcal{F}$ and evaluate accuracy of predicting the exact final version.
We then report the best score among the 10 seed representations as the performance metric on $\mathcal{F}$.

\rTab{tab:exp:transfer_learning} summarizes the results and also reports the upper bound performance 
when using the gold-standard edit representation \editrepr{\elementbefore}{\elementafter} to predict $\elementafter$,
and an approximation of the ``lower bound'' accuracies using pre-trained Seq2Seq and Graph2Tree models without edit encoders.
We found that our neural Graph2Tree editor with the sequential edit encoder significantly outperforms the Seq2Seq editor,
even though Seq2Seq performs better when using gold-standard edit representations.
This suggest that the edit representations learned with the Graph2Tree model generalize better,
especially for edits discussed in \rSC{sec:exp:edit_repr_quality} that involve syntactic variations like \texttt{RCS1021} 
(lambda expression simplification, 7.8\% \vs~30.7\% for Seq2Seq and Graph2Tree),
and \texttt{RCS1207} (change lambdas to method groups, 7.1\% \vs~26.2\%). 
Interestingly, we also observe that Seq2Seq outperforms the Graph2Tree model for edits with trivial surface edit sequences, where
the Graph2Tree model does not have a clear advantage.
For example, on \texttt{RCS1015} (use \texttt{nameof} operator, \eg \change{\code{Exception("x")}}{\code{Exception(nameof(x))}}),
the accuracies for Seq2Seq and Graph2Tree are 40.0\%  (14/35) and 28.6\% (10/35), resp.
We include more analysis of the results in~\autoref{app:zero_shot_exp}.

\begin{table}[tb]
	\footnotesize\centering
	\caption{Transfer learning results on C\# fixers data, averaged across all fixer categories.}
	\label{tab:exp:transfer_learning}
	\centering

	\begin{tabular}{lcccc}
	\toprule
	\textbf{Model} & \textbf{Acc.}(\%) & \textbf{Acc.}$^*$(\%) & \textbf{Recall@5}(\%) & \textbf{Recall@5}$^*$(\%) \\
	\midrule
	Seq2Seq -- Seq Edit Encoder & 38.35 & {\bf 77.67} & 41.50 & {\bf 83.84} \\
	Graph2Tree -- Seq Edit Encoder & {\bf 49.21} & 77.30 & {\bf 51.93} & 81.77 \\
  \hspace{-0.15cm}{\underline{Baselines} (no edit encoding)} \\
  Seq2Seq w/o Edit Encoder & 7.07 & --- & 14.29 & --- \\
  Graph2Tree w/o Edit Encoder & 8.81 & --- & 11.90 & --- \\
	\bottomrule
	\end{tabular}

	{\scriptsize  $^*$: upper-bound performance of predicting $\elementafter$ using the gold-standard edit representations.}
\end{table}

\vspace{-3mm}
\section{Related Work}
\vspace{-2mm}
\label{sect:rel_work}
%!TEX root = ../editrepresentations.tex

Edits have recently been considered in NLP,
as they represent interesting linguistic phenomena in 
language modeling and discourse~\citep{faruqui2018wikiatomicedits,yang17wiki}.
Specifically, \citet{guu2017generating} present a generative model of natural language sentences via editing prototypes.
%, where the generative story of a sentence is modeled by editing similar sentences from training data 
Our work shares with~\citet{guu2017generating} in that (1) the posterior edit encoding model in~\citet{guu2017generating} is similar to our baseline ``bag-of-edits'' encoder in Sec.~\ref{sec:perf_neural_editor}, and (2) the sequence-to-sequence sentence generation model given the prototype and edit representation is reminiscent of our Seq2Seq editor.
In contrast, our work directly focuses on discriminative learning of representing edits and  applying the learned edits for both sequential (NL) and structured (code) data.
Another similar line of research is ``retrieve-and-edit'' models for text generation~\citep{Hashimoto2018ARF}, where given an input data $\vect{x}$, the target prediction $\vect{y}$ is generated by editing a similar target $\vect{y}'$ that is retrieved based on the similarity between its source $\vect{x'}$ and the input $\vect{x}$.
While these models typically require an ``editor'' component to generate the output by exploiting the difference between similar inputs, they usually use the simpler bag-of-edits representations~\citep{Wu2018ResponseGB}, or implicitly capture it via end-to-end neural networks~\citep{Contractor2018ExemplarEF}.
To our best knowledge, there is not any related work that classifies
or otherwise explicitly represents the differences over similar input,
with the exception of differential recurrent neural networks
used for action recognition in videos~\citep{veeriah2015differential,zhuang2018deep}.
This is a substantially different task, as the data includes a
temporal component as well.

Source code edits are a widely studied artifact.
Specialized software, such as
\href{https://git-scm.com/}{git}, is widely used to store source
code revision histories.
\citet{nguyen2013study} studied the repetitiveness
of source code changes by identifying identical types of changes
using a deterministic differencing tool.
In contrast, we employ on a neural network to cluster \emph{similar}
changes together.
\citet{rolim2017learning} use such clusters to synthesize small programs that
perform the edit.
The approach is based on
\citet{rolim2018learning}
extract manually designed syntactic features from code
and cluster over multiple changes to find repeatable edit rules.
Similarly, \citet{paletov2018inferring} extract syntactic features
specifically targeting edits in cryptography API protocols. In this
work, we try to avoid hand-designed features
and allow a neural network to learn the relevant aspects of
a change by directly giving as input the original and final version
of a changed code snippet.

Modeling tree generation with machine learning is an old problem
that has been widely studied in NLP. Starting with \citet{maddison2014structured},
code generation has also been considered as a tree generation problem.
Close to our work is the decoder of \citet{yin2017syntactic} which
we use as the basis of our decoder. The work of \citet{chen2018tree}
is also related, since it provides a tree-to-tree model, but focuses
on learning a single translation tasks and cannot be used directly to represent multiple types of edits.
Both \citet{yin2017syntactic} and \citet{chen2018tree} have copying
mechanism for single tokens, but our subtree copying mechanism is novel.

Autoencoders (see \citet{goodfellow2016deep} for an overview) have a long history in machine
learning. Variational autoencoders~\citep{kingma2013auto} are similar to autoencoders but
instead of focusing on the learned representation, they aim to
create accurate generative probabilistic models.
Most (variational) autoencoders focus on encoding images
but there have been works that autoencode sequences,
such as text~\citep{dai2015semi,bowman2015generating,yang2017improved} and
graphs~\citep{simonovsky2018graphvae,liu2018constrained}.
Conditional variational autoencoders~\citep{sohn2015learning} have a
related form to our model (with the exception of the KL term), but
are studied as generative models, whereas we are primarily interested
in the edit representation.

\vspace{-3mm}
\section{Discussion \& Conclusions}
\vspace{-2mm}
In this work, we presented the problem of learning distributed
representation of edits.
We believe that the dataset of edits is highly relevant and should be
studied in more detail.
While we have presented a set of initial models and metrics on the problem
and obtained some first promising results, further development in both of
these areas is needed.
We hope that our work inspires others to work on this interesting problem
in the future.

\subsubsection*{Acknowledgments}
We would like to thank Rachel Free for her insightful comments and suggestions.

\bibliography{bibliography}
\bibliographystyle{iclr2019_conference}

\newpage
\appendix

\section{Datasets and Configuration}
\label{app:datasets}
%!TEX root = ../editrepresentations.tex

\paragraph{WikiAtomicEdits} We randomly sampled 1040$K$ insertion examples from the English portion of WikiAtomicEdits~\citep{faruqui2018wikiatomicedits} dataset, with a train, development and test splits of $1000K$, $20K$ and $20K$.

\paragraph{GitHubEdits} 
We cloned the top 54 C\# GitHub repositories based on their popularity (\rTab{tbl:csharpdataset}).
For each commit in the \texttt{master} branch, we collect the previous and updated versions of the source code, and extract all consecutive lines of edits that are smaller than three lines, and with at least three preceding and successive lines that have not been changed.
We then filter trivial changes such as variable and identifier renaming, and changes happened within comments.
We also limit the number of tokens for each edit to be smaller than 100, and down-sample edits whose frequency is larger than 30.
Finally, we split the dataset by commit ids, ensuring that there are no edits in the training and testing (development) sets coming from the same commit. \rTab{tab:app:github_data_stat} lists some statistics of the dataset. 

\begin{table}[H]
    \caption{Statistics of the GitHubEdits Dataset} \label{tab:app:github_data_stat}
    \centering
    \begin{tabular}{lr}
    \toprule
    Average Num. Tokens in \elementbefore & 16.4 \\
    Average Num. Tokens in \elementafter & 17.0 \\
    Average Edit Distance &  5.0  \\
    Average size of AST for $T(\elementbefore)$ & 28.5 \\
    Average size of AST for $T(\elementafter)$ & 29.4 \\
    \bottomrule
    \end{tabular}
\end{table}

\paragraph{C\#Fixers}
We selected 16 C\# fixers from Roslyn\footnote{\url{http://roslyn.io}} and Roslynator\footnote{\url{https://github.com/JosefPihrt/Roslynator}},  and ran them on 6 C\# projects to generate a small, high-quality C\# fixers dataset of 2\,878 edit pairs with known semantics. 
\autoref{tbl:csharp_fixer_data} lists the detailed descriptions for each fixer category. And more information can be found at \url{https://github.com/JosefPihrt/Roslynator/blob/master/src/Analyzers/README.md}.

\paragraph{Network Configuration} Throughout the experiments, we use a fixed edit representation size of 512.
The dimensionality of word embedding, the hidden states of the encoder LSTMs, as well as the gated graph neural network is 128, while the decoder LSTM uses a larger hidden size of 256.
For the graph-based edit encoder, we used a two-layer graph neural network, with 5 information propagation steps at each layer.
During training, we performed early stopping, and choose the best model based on perplexity scores on development set.
During testing, we decode a target element \elementafter using a beam size of 5.

\begin{table}
    \caption{Descriptions of fixer categories in C\#Fixers dataset} \footnotesize
    \label{tbl:csharp_fixer_data}
    \centering
    \begin{tabular}{lp{4.5cm}c@{}l}
    \toprule
    \textbf{Fixer ID} & \textbf{Description} & \textbf{Num. Edits} & ~~~~\textbf{Example}  \\
    \midrule
        \multirow{2}*{\texttt{CA2007}}  & \multirow{2}{5cm}{apply \code{.ConfigureAwait(false)} to \code{await} statements} &  \multirow{2}*{1051}  & \elementbefore: \lstinline|await Console.WriteAsync()| \\
         & & & \elementafter: \lstinline|await Console.WriteAsync()| \\
         & & & \hfill\lstinline|  .ConfigureAwait(false)| \\
    \midrule
    \multirow{2}*{\texttt{IDE0004}}  & \multirow{2}{5cm}{Cast is redundant} &  \multirow{2}*{53}  & \elementbefore: \lstinline|var x = 1; var b = (int)x;| \\
         & & & \elementafter: \lstinline|var x = 1; var b = x;| \\
    \midrule
    \multirow{2}*{\texttt{RCS1015}}  & \multirow{2}{5cm}{Use \code{nameof} operator} &  \multirow{2}*{35}  & \elementbefore: \lstinline|Exception("parameter");| \\
         & & & \elementafter: \lstinline|Exception(nameof(parameter));| \\
         \midrule
    \multirow{2}*{\texttt{RCS1021}}  & \multirow{2}{5cm}{Simplify lambda expression} &  \multirow{2}*{411}  & \elementbefore: \begin{lstlisting}
var x = items.Select(f =>
{
    return f.ToString();
});
\end{lstlisting} \\
         & & & \elementafter: \begin{lstlisting}
var x = items.Select(
        f => f.ToString());
\end{lstlisting} \\
    \midrule
    \multirow{2}*{\texttt{RCS1032}}  & \multirow{2}{5cm}{Remove redundant parentheses} &  \multirow{2}*{24}  & \elementbefore: \begin{lstlisting}
if ((x)) {}
\end{lstlisting} \\
         & & & \elementbefore: \lstinline|if (x) {}| \\
    \midrule
    \multirow{2}*{\texttt{RCS1058}}  & \multirow{2}{5cm}{Use compound assignment} &  \multirow{2}*{43}  & \elementbefore: \lstinline|i = i + 2;| \\
         & & & \elementafter: \lstinline|i += 2;| \\
    \midrule
    \multirow{2}*{\texttt{RCS1077}}  & \multirow{2}{8cm}{Optimize LINQ method call} &  \multirow{2}*{200}  & \elementbefore: \lstinline|items.Where(f => Foo(f)).Any();| \\
         & & & \elementafter: \lstinline|items.Any(f => Foo(f));| \\
    \midrule
    \multirow{2}*{\texttt{RCS1089}}  & \multirow{2}{5cm}{Use \code{--}/\code{++} operator instead of assignment} &  \multirow{2}*{75}  & \elementbefore: \lstinline|i = i + 1;| \\
         & & & \elementafter: \lstinline|i += 1;| \\
    \midrule
    \multirow{2}*{\texttt{RCS1097}}  & \multirow{2}{5cm}{Remove redundant \code{ToString} call} &  \multirow{2}*{20}  & \elementbefore: \lstinline|var x = s.ToString();| \\
         & & & \elementafter: \lstinline|var x = s;| \\
    \midrule
    \multirow{2}*{\texttt{RCS1118}}  & \multirow{2}{5cm}{Mark local variable as const} &  \multirow{2}*{477}  & \elementbefore: \begin{lstlisting}
string s = "a";
string s2 = s + "b";
\end{lstlisting} \\
         & & & \elementafter: \begin{lstlisting}
const string s = "a";
string s2 = s + "b";
\end{lstlisting} \\ \midrule
    \multirow{2}*{\texttt{RCS1123}}  & \multirow{2}{5cm}{Add parentheses according to operator precedence} &  \multirow{2}*{109}  & \elementbefore: \begin{lstlisting}
if (x || y && z) {}
\end{lstlisting} \\
         & & & \elementafter: \begin{lstlisting}
if (x || (y && z) ) {}
\end{lstlisting} \\
    \midrule
    \multirow{2}*{\texttt{RCS1146}}  & \multirow{2}{5cm}{Use conditional access} &  \multirow{2}*{71}  & \elementbefore: \lstinline|x != null && x.StartsWith("a");| \\
         & & & \elementafter: \lstinline|x?.StartsWith("a");| \\
    \midrule
    \multirow{3}*{\texttt{RCS1197}}  & \multirow{2}{5cm}{Optimize call of \code{StringBuilder}'s \code{Append}/\code{AppendLine}} &  \multirow{2}*{95}  & \elementbefore: \lstinline|sb.Append(s + "x");| \\
         & & & \elementafter: \lstinline|sb.Append(s).Append("x");| \\ \\
    \midrule
    \multirow{2}*{\texttt{RCS1202}}  & \multirow{2}{5cm}{Avoid \code{NullReferenceException}} &  \multirow{2}*{56}  & \elementbefore: \lstinline|items.First().ToString();| \\
         & & & \elementafter: \lstinline|items?.First().ToString();| \\
    \midrule
    \multirow{2}*{\texttt{RCS1206}}  & \multirow{2}{5cm}{Use conditional access instead of conditional expression} &  \multirow{2}*{116}  & \elementbefore: \begin{lstlisting}
int i = (x != null) ? 
        x.Value.GetHashCode() : 0;
\end{lstlisting} \\
         & & & \elementafter: \begin{lstlisting}
int i = x?.GetHashCode() ?? 0;
\end{lstlisting} \\ \midrule
\multirow{2}*{\texttt{RCS1207}}  & \multirow{2}{5cm}{Use method group instead of anonymous function} &  \multirow{2}*{42}  & \elementbefore: \lstinline|items.Select(f => Foo(f));| \\
         & & & \elementafter: \lstinline|items.Select(Foo);| \\
    \bottomrule
    \end{tabular}
\end{table}

\begin{table}
    \caption{Our C\# GitHub dataset projects}\label{tbl:csharpdataset}
    \scriptsize\centering
\begin{tabular}{lp{4cm}p{4.5cm}}
    \toprule
    Name & GitHub Id & Description \\
    \midrule
acat & intel/acat & Assistive Context-Aware Toolkit  \\
akka.net & akka/akka.net & Distributed Actors \\
aspnetboilerplate & aspnetboilerplate/aspnetboilerplate & ASP.NET boilerplate\\
AutoMapper & AutoMaper/AutoMapper & Object-Object Mapper\\
BotBuilder & Microsoft/BotBuilder & Bot Framework\\
CefSharp & cefsharp/CefSharp & Chromium Embedded Framework Bindings \\
choco & chocolatey/choco & package mananger \\
cli & dotnet/cli & .NET CLI Tools \\
CodeHub & CodeHubApp/CodeHub & iOS application\\
coreclr & dotnet/coreclr & .NET Framework\\
corefx & dotnet/corefx & .NET FOundational Libraries \\
dapper & StackExchange/Dapper & Object Mapper\\
dnSpy & 0xd4d/dnSpy & .NET debugger and assembly editor \\
duplicati & duplicati/duplicati & Encrypted Cloud Backups \\
EntityFramework & aspnet/EntityFramework & Object-Relational Mapper\\
EntityFrameworkCore & aspnet/EntityFrameworkCore & Object-Relational Mapper -- Core\\
FluentValidation & JeremySkinner/FluentValidation & Validation Rules \\
framework & accord-net/framework &ML, CV Framework \\
GVFS & Microsoft/VFSForGit & Git Virual File System \\
Hangfire & HangfireIO/Hangfire & Background job library \\
ILSpy & icsharpcode/ILSpy & Decompiler \\
JavaScriptServices & aspnet/JavaScriptServices & ASP.NET JS Services\\
MahApps.Metro & MahApps/MahApps.Metro & WPF Framework\\
MaterialDesignInXamlToolkit & MaterialDesignInXamlToolkit/ MaterialDesignInXamlToolkit & Design XAML \& WPF\\
mono & mono/mono & .NET implementation\\
monodevelop & mono/monodevelop & IDE\\
MonoGame & MonoGame/MonoGame & Game Framework \\
msbuild & Microsoft/msbuild & Build Tool\\
Mvc & aspnet/Mvc & MVC Framework\\
Nancy & NancyFx/Nancy & HTTP based services \\
Newtonsoft.Json & JamesNK/Newtonsoft.Json & JSON framework\\
NLog & NLog/NLog &  Loggin for .NET\\
OpenLiveWriter & OpenLiveWriter/ OpenLiveWriter & Text editor \\
OpenRA & OpenRA/OpenRA & Strategy Game Engine\\
Opserver & opserver/Opserver & Monitoring System\\
orleans & dotnet/orleans & Distributed Virtual Actors\\
PowerShell & PowerShell/PowerShell & Command Line\\
Psychson & brandonlw/Psychson & Firmware\\
PushSharp & Redth/PushSharp & Push Notifications\\
ravendb & ravendb/ravendb & Database \\
ReactiveUI & reactiveui/ReactiveUI & Reactive MVC Framework \\
RestSharp & restsharp/RestSharp & HTTP/REST Client \\
roslyn & dotnet/roslyn & .NET Compiler \\
Rx.NET & dotnet/reactive & Reactive extensions.\\
ServiceStack & ServiceStack/ServiceStack & Web Service Framework\\
shadowsocks-windows & shadowsocks/ shadowsocks-windows & Cryptography\\
ShareX & ShareX/ShareX & Screen Recorder\\
SignalR & SignalR/SignalR & Real-time web framework\\
Sonarr & Sonarr/Sonarr & PVR \\
SpaceEngineers & KeenSoftwareHouse/ SpaceEngineers & Game \\
SparkleShare & hbons/SparkleShare & File Sharing\\
StackExchange.Redis & StackExchange/ StackExchange.Redis & Redis Client\\
WaveFunctionCollapse & mxgmn/ WaveFunctionCollapse & Bitmap/tilemap Generator\\
Wox & Wox-launcher/Wox & Launcher\\ \bottomrule
\end{tabular}
\end{table}

\newpage
\section{Clustering Experiments}
\label{app:clustering_exp}
%!TEX root = ../editrepresentations.tex

To qualitatively evaluate the quality of the learned edit representations. 
We use the models trained on the WikiAtomicEdits and GitHubEdits datasets to cluster natural language and code edits.
We run K-Means clustering algorithm on $0.5$ million sampled edits from WikiAtomicEdits, and all $90K$ code edits from GitHubEdits, producing 50\,000 and 20\,000 clusters for each dataset.

\rTab{tbl:wiki_data_clusters} and \rTab{tbl:github_data_clusters} list some example clusters on WikiAtomicEdits and GitHub datasets, respectively.
Due to the size of clusters, we omit out-liners and present distinctive examples from each cluster.
On the WikiAtomicEdits dataset, we found clusters whose examples are semantically and syntactically similar.
More interestingly, on the source code data, we find representative clusters that relate to idiomatic patterns and best practices of programming.
The clustering results produced by our model would be useful for programming synthesis toolkits to generate interpretable code refractory rules, which we leave as interesting future work.

Finally, we remark that the clustering results indicate that the encoding of edits is \emph{context-sensitive} and \emph{position-sensitive} for both natural language and source code data.
For instance, the WikiAtomicEdits examples we present in \rTab{tbl:wiki_data_clusters} clearly indicate that semantically similar insertions also share similar editing positions.
This is even more visible in code edits (\rTab{tbl:github_data_clusters}). For instance, in the first example in \rTab{tbl:github_data_clusters}, \code{Equal()} can be changed to \code{Empty()} only in the \code{Assert} namespace (i.e., the context).
These examples demonstrate that it is important for an edit encoder to capture the contextual and positional information in edits, a property that cannot be captured by simple ``bag-of-edits'' edit representation methods.

\begin{table}
\caption{Example clusters on WikiAtomicEdits data using representations learned by a neural Seq2Seq editor with sequential edit encoder}\label{tbl:wiki_data_clusters}
\footnotesize
\centering
\begin{tabular}{p{1mm}p{12cm}} \toprule
	\multicolumn{2}{l}{\underline{Description}~~Add a person's middle name} \\ \rowspace
    1. & isaiah \insertText{marcus} rankin ( born 22 may 1978 in london ) is an english professional footballer currently playing for stevenage borough .\\

    2. & audrey \insertText{kathleen} brown ( born 24 may , 1913 ) is a british athlete who competed mainly in the 100 metres . \\

    3. & alice \insertText{edith} rumph was a painter , etcher , and teacher .\\

    4. & mark \insertText{larry} taufua is an australian professional rugby league player .\\

    5. & monique \insertText{edith} lamoureux ( born july 3 , 1989 ) is an american ice hockey player .\\

    \midrule 

	\multicolumn{2}{l}{\underline{Description}~~Add a parenthetical expression \textit{also ... as} to modify the subject} \\ \rowspace
    1. & mid-state regional airport \insertText{, also known as mid-state airport ,} is a small airport on in rush township , centre county in pennsylvania in the united states .\\

    2. & islamic culture \insertText{, also known as saracenic culture ,} is a term primarily used in secular academia to describe the cultural practices common to historically islamic peoples . \\

    3. & birds of prey \insertText{, also known as raptors ,} are birds that hunt for food primarily via flight , using their keen senses , especially vision .\\

    4. & tetyana styazhkina \insertText{, also written as tetyana stiajkina ,} ( ; born april 10 , 1977 ) is a ukrainian cycle racer who rides for the chirio forno d'asolo team .\\

    5. & acid jazz \insertText{, also known as club jazz ,} is a musical genre that combines elements of jazz , soul , funk , disco and hip hop .\\
    \midrule 

    \multicolumn{2}{l}{\underline{Description}~~Specify location using a prepositional phrase.} \\ \rowspace
    1. &  the douro fully enters portuguese territory just after the confluence with the águeda river ; once the douro enters portugal , major           population centres are less frequent \insertText{along the river} .\\

    2. & mochou lake and mochou lake park are located at 35 hanzhongmen da jie in the jianye district of nanjing , china \insertText{, west to qinhuai river} . \\

    3. & reiner gamma is an albedo feature that is located on the oceanus procellarum , to the west of the reiner crater \insertText{on the moon} .\\

    4. & she made a brief return to the screen in '' parrish '' ( 1961 ) , playing the supporting role of mother which received little attention \insertText{by the press} .\\

    5. & he was involved in a few storylines , including one where he broke his toe and had a heart attack after he was pushed by a mugger \insertText{in the market} .\\ 
    \midrule

    \multicolumn{2}{l}{\underline{Description}~~Add positional or temporal clause} 
    \\ \rowspace
    1. &~\insertText{at the time} ajax and hercules were trapped behind a landslide at the gaillard cut , both were working to clear the landslide . \\

    2. &~\insertText{at the docks ,} hikaru attempts to befriend the tiger , but finds that it dislikes humans . \\

    3. &~\insertText{about the second ,} i do know they exist , but the question is whether they are considered a genre outside of japan .\\

    4. &~\insertText{in the battle ,} shirou uses his reality marble , unlimited blade works and defeats gilgamesh .\\

    5. &~\insertText{in the game ,} red is a curious 11 - year - old boy from pallet town . \\

    \bottomrule
\end{tabular}   
\end{table}

%<<<<<<< HEAD
\begin{table}
\caption{Example clusters on GithubEdits data using representations learned by a Graph2Tree editor with sequential edit encoder. Locally defined variable names are canonicalized.}\label{tbl:github_data_clusters}
% =======
% \begin{table}[h]
% \caption{Example clusters on GitHubEdits data using representations learned by a Graph2Tree editor with sequential edit encoder. Locally defined variable names are canonicalized.}\label{tbl:github_data_clusters}
% >>>>>>> origin/arxiv
\footnotesize
\centering
\begin{tabular}{ll} \toprule
	\multicolumn{2}{l}{\underline{Description}~~Switch from \code{Assert.Equal} to \code{Assert.Empty}} \\ \rowspace
    \elementbefore & \begin{lstlisting}
Assert.Equal(0, V0.ProjectIds.Count);
\end{lstlisting} \\
	\elementafter & \begin{lstlisting}
Assert.Empty(V0.ProjectIds);
\end{lstlisting} \\ \rowspace

    \elementbefore & \begin{lstlisting}
Assert.Equal(0, V0.ProjectReferences.Count());
\end{lstlisting} \\
	\elementafter & \begin{lstlisting}
Assert.Empty(V0.ProjectReferences);
\end{lstlisting} \\ \rowspace

    \elementbefore & \begin{lstlisting}
Assert.Equal(0, V0.TrustedSelectionPaths.Count);
\end{lstlisting} \\
	\elementafter & \begin{lstlisting}
Assert.Empty(V0.TrustedSelectionPaths);
\end{lstlisting} \\ \rowspace

    \elementbefore & \begin{lstlisting}
Assert.Equal(0, V0.Count);
\end{lstlisting} \\
	\elementafter & \begin{lstlisting}
Assert.Empty(V0);
\end{lstlisting} \\ \rowspace

    \elementbefore & \begin{lstlisting}
Assert.Equal(0, V0.Messages.Count);
\end{lstlisting} \\
	\elementafter & \begin{lstlisting}
Assert.Empty(V0.Messages);
\end{lstlisting} \\ \midrule

	\multicolumn{2}{l}{\underline{Description}~~Use conditional access} \\ \rowspace
    \elementbefore & \begin{lstlisting}
Type V0 = V1 == null ? null : V1.GetType();
\end{lstlisting} \\
	\elementafter & \begin{lstlisting}
Type V0 = V1?.GetType();
\end{lstlisting} \\ \rowspace

    \elementbefore & \begin{lstlisting}
V0 = ((V1!= null) ? V1.Operand : null) as MemberExpression;
\end{lstlisting} \\
	\elementafter & \begin{lstlisting}
V0 = V1?.Operand as MemberExpression;
\end{lstlisting} \\ \rowspace

    \elementbefore & \begin{lstlisting}
string V0 = V1 == null ? null : V1.GetType().Name;
\end{lstlisting} \\
	\elementafter & \begin{lstlisting}
string V0 = V1?.GetType().Name;
\end{lstlisting} \\ \rowspace

    \elementbefore & \begin{lstlisting}
var V0 = V1 == null ? null : V1(V2).ToArray();
\end{lstlisting} \\
	\elementafter & \begin{lstlisting}
var V0 = V1?.Invoke(V2).ToArray();
\end{lstlisting} \\ \midrule

	\multicolumn{2}{l}{\underline{Description}~~Optimize LINQ queries} \\ \rowspace
    \elementbefore & \begin{lstlisting}
var V0 = V1.Customers.Where(V2 => V2.CustomerID == LITERAL)
                    .FirstOrDefault();
\end{lstlisting} \\
	\elementafter & \begin{lstlisting}
var V0 = V1.Customers
               .FirstOrDefault(V2 => V2.CustomerID == LITERAL);
\end{lstlisting} \\ \rowspace

    \elementbefore & \begin{lstlisting}
var V0 = V1.TypeConverters.Where(V2 => V2.CanConvertTo(V3, V1))
               .FirstOrDefault();\end{lstlisting} \\
	\elementafter & \begin{lstlisting}
var V0 = V1.TypeConverters
              .FirstOrDefault(V2 => V2.CanConvertTo(V3, V1));
\end{lstlisting} \\ \rowspace

    \elementbefore & \begin{lstlisting}
var V0 = this.V1.Where(V2 => V2.CanDeserialize(V3))
					            .FirstOrDefault();
\end{lstlisting} \\
	\elementafter & \begin{lstlisting}
var V0 = this.V1.FirstOrDefault(V2 => V2.CanDeserialize(V3));
\end{lstlisting} \\ \rowspace

    \elementbefore & \begin{lstlisting}
var V0 = V1.Where(V2 => V2.Item1 == V3 && V2.Item2 == V4)
                    .FirstOrDefault();
\end{lstlisting} \\
	\elementafter & \begin{lstlisting}
var V0 = V1.FirstOrDefault(V2 => V2.Item1 == V3 && V2.Item2 == V4);
\end{lstlisting} \\ \midrule   

\multicolumn{2}{l}{\underline{Description}~~Change from \code{Add} function to indexer.} \\ \rowspace
    \elementbefore & \begin{lstlisting}
V0.Add(V1.key, V1.V2);
\end{lstlisting} \\
	\elementafter & \begin{lstlisting}
V0[V1.key] = V1.V2;
\end{lstlisting} \\ \rowspace

    \elementbefore & \begin{lstlisting}
V0.Add(V1.Id, V2);
\end{lstlisting} \\
	\elementafter & \begin{lstlisting}
V0[V1.Id] = V2;
\end{lstlisting} \\ \rowspace

    \elementbefore & \begin{lstlisting}
V0.Add(V1.Etag, V1);
\end{lstlisting} \\
	\elementafter & \begin{lstlisting}
V0[V1.Etag] = V1;
\end{lstlisting} \\ \rowspace

    \elementbefore & \begin{lstlisting}
V0.Add(V1.V2, V3.Merge(V1.V4));
\end{lstlisting} \\
	\elementafter & \begin{lstlisting}
V0[V1.V2] = V3.Merge(V1.V4);
\end{lstlisting} \\

    \bottomrule
\end{tabular}   
\end{table}

\newpage
\newpage
\newpage
\section{Break-down Analysis of Transfer Learning Results}
\label{app:zero_shot_exp}
%!TEX root = ../editrepresentations.tex

\rTab{tbl:zero_shot_exp_break_down} lists the detailed evaluation results for the transfer learning experiments discussed in \rSC{sec:perf_neural_editor}. We refer readers to \rTab{tbl:csharp_fixer_data} for detailed descriptions of each fixer category.
The neural Graph2Tree editor outperforms the Seq2Seq editor (both with sequential edit encoders) on 10 out of 16 fixer categories.
However, we found that there are categories where both end-to-end system under-performs, even though the upper-bound accuracy is high (\eg \texttt{RCS1077}, \texttt{RCSRCS1197}, \texttt{RCS1207}, \texttt{RCS1032}). 
While improving the generalization ability of the neural editor models to achieve better transfer learning performance is an important future work, we remark that this task is indeed non-trivial. 
First, some fixer categories cover a broad range of similar edits, which could not be captured by a single seed edit.
xSecond, some categories contain syntactically or semantically complex refactoring rules.
For instance, \texttt{RCS1207} converts method groups into anonymous functions, involving changing multiple positions of the source code, which might not be trivially captured by the sequential edit encoder from a single example edit.
Additionally, \texttt{RCS1197} requires reasoning about a chain of expressions. It turns \code{sb.Append(s1 + s2 + $\ldots$ + sN)} into \code{sb.Append(s1).Append(s2).[$\ldots$]Append(sN))}, which our current models are unable to reason about.
More interestingly, we found that there are cases where the edits are syntactically simple, but could be semantically more difficult to learn.
For instance, \texttt{RCS1032} is about removing redundant parentheses from expressions.
Although the edit pattern might seem to be syntactically simple at the AST level (replacing a \code{ParethesizedExpressionSyntax} node by its child node), determining which pair of parentheses is actually redundant in an expression (\eg \code{(a + b) * (c / d)}) is semantically non-trivial to learn from a single edit example.   
We believe that further advances in (general) learning from source code are required to correctly handle these cases.

%While these are more complex refractory rules where a single local seed edit might not be able to generalize to all scenarios, improving the generalization ability of the edit representations and the performance of end-to-end system would be interesting future directions. 

\begin{table}[tb]
	\caption{Break-down performance results on the transfer learning task. See \rTab{tbl:csharp_fixer_data} for descriptions of each fixer category.}
	\label{tbl:zero_shot_exp_break_down}
	\scriptsize
	\centering

	\begin{tabular}{lrrrr|rrrr}
	\hline

	\hline
	 & \multicolumn{4}{c}{Graph2Tree --- Seq Edit Encoder} & \multicolumn{4}{c}{Seq2Seq --- Seq Edit Encoder} \\
	\textbf{Fixer ID} & \textbf{Acc.}(\%) & \textbf{Acc.}$^*$(\%) & \textbf{Recall@5}(\%) & \textbf{Recall@5}$^*$(\%) & \textbf{Acc.}(\%) & \textbf{Acc.}$^*$(\%) & \textbf{Recall@5}(\%) & \textbf{Recall@5}$^*$(\%) \\
	\hline
	\texttt{CA2007} & 88.0 & 89.2 & 88.2 & 94.3 & 52.7 & 91.9 & 61.0 & 93.8 \\
	\texttt{IDE0004} & 69.8 & 92.5 & 73.6 & 94.3 & 45.3 & 98.1 & 45.3 & 98.1 \\
	\texttt{RCS1015} & 28.6 & 82.9 & 40.0 & 82.9 & 40.0 & 71.4 & 42.9 & 71.4 \\
	\texttt{RCS1021} & 30.7 & 60.8 & 33.3 & 67.6 & 7.8 & 56.2 & 17.8 & 72.3 \\
	\texttt{RCS1032} & 8.3 & 37.5 & 8.3 & 45.8 & 20.8 & 45.8 & 20.8 & 45.8 \\
	\texttt{RCS1058} & 93.0 & 88.4 & 95.3 & 90.7 & 37.2 & 69.8 & 39.5 & 76.7 \\
	\texttt{RCS1077} & 6.5 & 69.5 & 6.5 & 74.0 & 7.5 & 84.0 & 7.5 & 84.5 \\
	\texttt{RCS1089} & 96.0 & 98.7 & 98.7 & 98.7 & 76.0 & 98.7 & 76.0 & 98.7 \\
	\texttt{RCS1097} & 15.0 & 90.0 & 15.0 & 90.0 & 25.0 & 90.0 & 25.0 & 95.0 \\
	\texttt{RCS1118} & 95.4 & 98.1 & 99.6 & 99.6 & 93.7 & 99.6 & 98.7 & 1.00 \\
	\texttt{RCS1123} & 66.1 & 81.7 & 68.8 & 86.2 & 64.2 & 87.2 & 65.1 & 94.5 \\
	\texttt{RCS1146} & 54.9 & 81.7 & 56.3 & 85.9 & 45.1 & 76.1 & 57.7 & 91.5 \\
	\texttt{RCS1197} & 5.3 & 25.3 & 5.3 & 33.7 & 12.6 & 40.0 & 12.6 & 50.0 \\
	\texttt{RCS1202} & 28.6 & 67.9 & 37.5 & 75.0 & 28.6 & 69.6 & 32.1 & 80.4 \\
	\texttt{RCS1206} & 75.0 & 99.1 & 75.9 & 99.1 & 50.0 & 1.00 & 50.0 & 1.00 \\
	\texttt{RCS1207} & 26.2 & 73.8 & 28.6 & 90.5 & 7.1 & 64.3 & 11.9 & 88.1 \\
	\hline

	\hline
	\end{tabular}

	{\scriptsize  $^*$: upper-bound performance of predicting $\elementafter$ using the gold-standard edit representations.}
\end{table}

\section{Impact of Training Set Size}
%!TEX root = ../editrepresentations.tex

To evaluate the data efficiency of our proposed approach, we tested the end-to-end performance of our neural editor model~(\rSC{sec:perf_neural_editor},~\rTab{tab:exp:neural_editor_precision}) with varying amount of training data. \rTab{tab:app:neural_editor_precision_vary_amount_training_data} lists the results.
We found both Graph2Tree and Seq2Seq editors are relatively data efficient.
They registered around $90\%$ of the accuracies achieved using the full training set with only $60\%$ of the training data.

\begin{table}[tb]
    \caption{Test performance of end-to-end experiments with varying amount of training data.}
    \label{tab:app:neural_editor_precision_vary_amount_training_data}
    \centering \footnotesize

    \begin{tabular}{llrrr}
    \toprule
    &\textbf{Training Set Size} & \textbf{Acc.@1} (\%) & \textbf{Recall@5} (\%) & \textbf{PPL} per token \\
    \midrule
        \multicolumn{3}{l}{\underline{GitHubEdits}} \\
        \multicolumn{3}{l}{Graph2Tree -- Seq Edit Encoder} \\
        & $20\%$ & 43.88 & 50.53 & 1.5703 \\
        & $40\%$ & 50.44 & 56.63 & 1.4152 \\
        & $60\%$ & 53.78 & 60.00 & 1.3720 \\
        & $80\%$ & 55.51 & 60.85 & 1.3392 \\
        & $100\%$ & {\bf 57.49} & {\bf 62.94} & {\bf 1.3043} \\
        \multicolumn{3}{l}{\underline{WikiAtomicEdits}} \\
        \multicolumn{3}{l}{Seq2Seq -- Seq Edit Encoder} \\
        & $20\%$ & 42.87 & 48.24 & 1.4123 \\
        & $40\%$ & 57.72 & 62.31 & 1.1812 \\
        & $60\%$ & 65.22 & 69.62 & 1.1070 \\
        & $80\%$ & 68.44 & 73.34 & 1.0751 \\
        & $100\%$ & {\bf 72.94} & {\bf 76.53} & {\bf 1.0527} \\
    \bottomrule
    \end{tabular}
\end{table}

\section{Details of Human Evaluation}
\label{app:annotation}
%!TEX root = ../editrepresentations.tex

As discussed in~\rSC{sec:exp:edit_repr_quality}, we performed human evaluation to rate the qualities of neighboring edits given a seed edit. 
The annotation instructions on GithubEdits and WikiAtomicEdits datasets are listed below. 
The annotation was carried out by three authors of this paper, and we anonymized the source of systems that generated the output. 
The three-way Fleiss' kappa inter-rater agreement is $\kappa=0.55$, which shows moderate agreement \citep{Artstein2008InterCoderAF}, an agreement level that is also used in other annotation tasks in NLP \citep{FaruquiDas2018}.

\begin{longtable}{p{\columnwidth}}
\caption{Annotation Instruction for GitHubEdits Data} \\

 \toprule
	\textbf{Rating 2} Semantically and Syntactically Equivalent \\ \midrule
	
	\emph{
		The changed constituents in the seed edit and the neighboring edit are applied to the similar positions of the original sentence, serving the same syntactic and semantic role. For example,
	}\\ \hline
	\textbf{Examples} \\
	\begin{itemize}
		\item \textbf{Seed Edit}

		\begin{tabular}{ll} 
    		\elementbefore & \lstinline|var V0 = V1.Where(V2 => V2.Name == LITERAL).Single();| \\
			\elementafter & \lstinline|var V0 = V1.Single(V2=> V2.Name == LITERAL);|\\
		\end{tabular}

		\item \textbf{Neighbor} 

		\begin{tabular}{ll} 
    		\elementbefore & \lstinline|var V0 = V1.GetMembers().Where(V2 => V2.Kind ==| \\
    		& \hspace{7cm} \lstinline|SymbolKind.Property).Single();| \\
			\elementafter & \lstinline|var V0 = V1.GetMembers().Single(V2 => V2.Kind ==| \\
			& \hspace{7cm} \lstinline|SymbolKind.Property);| \\
		\end{tabular}
	\end{itemize}

	\begin{itemize}
		\item \textbf{Seed Edit}
		
		\begin{tabular}{ll} 
    		\elementbefore & \lstinline|Type V0 = V1 == null ? typeof(object) : V1.GetType();| \\
			\elementafter & \lstinline|Type V0 = V1?.GetType() ?? typeof(object);| \\
		\end{tabular}

		\item \textbf{Neighbor} 

		\begin{tabular}{ll} 
    		\elementbefore & \lstinline|string V0 = V1 == null ? string.Empty : VAR1.ToString();| \\
			\elementafter & \lstinline|string V0 = V1?.ToString() ?? string.Empty;| \\
		\end{tabular}
	\end{itemize}

	\begin{itemize}
		\item \textbf{Seed Edit}
		
		\begin{tabular}{ll} 
    		\elementbefore & \lstinline|Assert.True(Directory.Exists(V0) == V1);| \\
			\elementafter & \lstinline|Assert.Equal(Directory.Exists(V0), V1);| \\
		\end{tabular}

		\item \textbf{Neighbor} 

		\begin{tabular}{ll} 
    		\elementbefore & \lstinline|Assert.True(V0.GetString(V0.GetBytes(LITERAL)) ==| \\
    		& \hspace{3cm} \lstinline|V1.ContainingAssembly.Identity.CultureName);| \\
			\elementafter & \lstinline|Assert.Equal(V0.GetString(VAR0.GetBytes(LITERAL)),| \\
			& \hspace{3cm} \lstinline|V1.ContainingAssembly.Identity.CultureName);| \\
		\end{tabular}
	\end{itemize} \\ 	

	\toprule

	\textbf{Rating 1} Syntactically or Semantically Related \\ \midrule
	\emph{
		The seed and neighboring edits share functionally or syntactically similar patterns.
	} \\ \hline
	\textbf{Examples} \\
	The following edit is a related edit of the first example above, as it applies the same simplification (\code{.Where(COND).Func()} to \code{.Func(COND)}), but for \code{FirstOrDefault} instead of \code{Single}:

	\begin{itemize}
		\item \textbf{Seed Edit}

		\begin{tabular}{ll} 
    		\elementbefore & \lstinline|var V0 = V1.Where(V2 => V2.Name == LITERAL).Single();| \\
			\elementafter & \lstinline|var V0 = V1.Single(V2=> V2.Name == LITERAL);|\\
		\end{tabular}

		\item \textbf{Neighbor} 

		\begin{tabular}{ll} 
    		\elementbefore & \lstinline|var V0 = V1.Where(V2 => V3.ReportsTo == V2.EmployeeID).FirstOrDefault();| \\
			\elementafter & \lstinline|var V0 = V1.FirstOrDefault(V2 => V3.ReportsTo == V2.EmployeeID);| \\
		\end{tabular}
	\end{itemize}

	The following edit is a related edit of the second example above, as it also replaces a ternary expression for null checking with the \code{?.} and \code{??} operators:

	\begin{itemize}
		\item \textbf{Seed Edit}

		\begin{tabular}{ll} 
    		\elementbefore & \lstinline|Type V0 = V1 == null ? typeof(object) : V1.GetType();| \\
			\elementafter & \lstinline|Type V0 = V1?.GetType() ?? typeof(object);| \\
		\end{tabular}

		\item \textbf{Neighbor} 

		\begin{tabular}{ll} 
    		\elementbefore & \lstinline|var V0 = V1 != null ? V1.ToList() : new List<TextSpan>();| \\
			\elementafter & \lstinline|var V0 = V1?.ToList() ?? new List<TextSpan>();| \\
		\end{tabular}
	\end{itemize}

	We also considered pairs such as the following related, since they share similar syntactic structure

	\begin{itemize}
		\item \textbf{Seed Edit}

		\begin{tabular}{ll} 
    		\elementbefore & \lstinline|V0.State = V1;| \\
			\elementafter & \lstinline|V0.SetState(VAR1);| \\
		\end{tabular}

		\item \textbf{Neighbor} 

		\begin{tabular}{ll} 
    		\elementbefore & \lstinline|V0.Quantity = V1;| \\
			\elementafter & \lstinline|V0.SetQuantity(V1);| \\
		\end{tabular}
	\end{itemize}
	\\ \toprule

	\textbf{Rating 0} Not Related \\ \midrule
	\emph{
		The seed and neighboring edits are not related based on the above criteria.
	} \\ \bottomrule
\end{longtable}

\begin{table}
\caption{Annotation Instruction for WikiAtomEdits Data}
\begin{tabular}{p{0.07\columnwidth}p{0.45\columnwidth}p{0.45\columnwidth}} \toprule
	\multicolumn{3}{p{\columnwidth}}{\textbf{Rating 2} Semantically and Syntactically Equivalent} \\ \midrule
	
	\multicolumn{3}{p{\columnwidth}}{\emph{
		The changed constituents in the seed edit and the neighboring edit are applied to the similar positions of the original sentence, serving the same syntactic and semantic role. For example,
	}} \\ \cmidrule{2-3}
	& Seed Edit  &  Neighbor \\ \cmidrule{2-3}

	&chaz guest ( born \insertText{1961}) was born in niagra falls , $\ldots$ , a decorated hero in wwii in europe , including the purple heart . &
	randal l. schwartz ( born november 22 , \insertText{1961}) , also known as merlyn , is an american author , system administrator and programming consultant. \\ \\

	&he was elected to donegal county council for sinn féin in 1979 , and held his seat until his death \insertText{at age 56} . &
	davis graduated from high school in january 1947 , immediately enrolling at wittenberg college in rural ohio \insertText{at age 17} . \\ \\

	&\insertText{dror} feiler served as a paratrooper in the israel defense forces . &
	\insertText{nagaur} fort - sandy fort ; centrally located ; 2nd century old ; witnessed many battles ; lofty walls \& spacious campus ; having many palaces \& temples inside . \\ \\

	&the original old bay house , home of the chief factor , still exists \insertText{and is now part of the fort vermilion national historic site} . &
	the population was 6,400 at the 2010 census \insertText{and is part of the st. louis metropolitan area} . \\  \toprule

	\multicolumn{3}{p{\columnwidth}}{\textbf{Rating 1} Syntactically Related} \\ \midrule
	\multicolumn{3}{p{\columnwidth}}{\emph{
		The changed constituents in the seed and the neighboring edit are applied to the similar positions of the original sentence, and they play similar syntactic roles. This includes examples like adding a disfunction, adding a complement, prepositional clause or other syntactic constructs with similar phrases or language structures. For example,
	}} \\ \cmidrule{2-3}
	&Seed Edit  &  Neighbor \\ \cmidrule{2-3}

	&the douro fully enters portuguese territory just after the confluence with the águeda river ; once the douro enters portugal , major population centres are less frequent \insertText{along the river} . &
	she made a brief return to the screen in '' parrish '' ( 1961 ) , playing the supporting role of mother which received little attention \insertText{by the press} . \\ \\

	&when they found it , they discovered a group of pagumon living there instead who immediately proceeded to treat the digidestined as honored guests \insertText{, saying that pagumon are the fresh form of koromon} . &
	in 2012 slote and his baseball book '' jake '' were the subject of an espn ( 30 for 30 ) short documentary in which slote describes his writing process and reads from the book \insertText{, saying it is his best writing} . \\ \\

	&the aircraft was intended to be \insertText{certified and} supplied as a complete ready - to - fly - aircraft for the flight training and aerial work markets . &
	in june reinforcements finally did arrive when \insertText{provincial and} militia units from new york , new jersey , and new hampshire were sent up from fort edward by general daniel webb . \\ \\ \toprule

	\multicolumn{3}{p{\columnwidth}}{\textbf{Rating 0} Not Related} \\ \midrule
	\multicolumn{3}{p{\columnwidth}}{\emph{
		The seed and neighboring edits are not related based on the above criteria.
	}}\\ \bottomrule
\end{tabular}
\end{table}

\end{document}